\pdfoutput=1

\documentclass[11pt]{article}

\usepackage{acl}

\usepackage{times}
\usepackage{latexsym}
\usepackage{booktabs}
\usepackage{multirow}
\usepackage{combelow}
\usepackage{tabularx}
\usepackage{xcolor}
\usepackage{lipsum}

\usepackage[T1]{fontenc}

\usepackage[utf8]{inputenc}

\usepackage{microtype}

\usepackage{inconsolata}

\usepackage{graphicx}

\usepackage{listings}

\lstnewenvironment{queryl}[1][] 
 {\lstset{frame=shadowbox,escapechar=`,linewidth=8cm, #1}}
 {}

\title{RoLargeSum: A Large Dialect-Aware Romanian News Dataset for Summary, Headline, and Keyword Generation}

\author{Andrei-Marius Avram$^{1}$, Mircea Timpuriu$^{1}$, Andreea Iuga$^{1}$, Vlad-Cristian Matei$^{1}$, \\ \textbf{Iulian-Marius Tăiatu$^{1}$, Tudor Găină$^{1}$, Dumitru-Clementin Cercel$^{1}$\thanks{Corresponding author: dumitru.cercel@upb.ro.}, Florin Pop$^{1,4}$,} \\
\textbf{Mihaela-Claudia Cercel$^{2,3}$} \\
$^{1}$  National University of Science and Technology POLITEHNICA Bucharest, Romania \\
$^{2}$ Paris 1 Panthéon-Sorbonne University, Paris, France \\
$^{3}$ University of Bucharest, Bucharest, Romania \\
$^{4}$ National Institute for Research and Development in Informatics - ICI Bucharest, Romania \\
}

\begin{document}
\maketitle
\begin{abstract}
Using supervised automatic summarisation methods requires sufficient corpora that include pairs of documents and their summaries. Similarly to many tasks in natural language processing, most of the datasets available for summarization are in English, posing challenges for developing summarization models in other languages. Thus, in this work, we introduce RoLargeSum, a novel large-scale summarization dataset for the Romanian language crawled from various publicly available news websites from Romania and the Republic of Moldova that were thoroughly cleaned to ensure a high-quality standard. RoLargeSum contains more than 615K news articles, together with their summaries, as well as their headlines, keywords, dialect, and other metadata that we found on the targeted websites. We further evaluated the performance of several BART variants and open-source large language models on RoLargeSum for benchmarking purposes. We manually evaluated the results of the best-performing system to gain insight into the potential pitfalls of this data set and future development.

\end{abstract}

\section{Introduction}

Text summarization, an essential task in natural language processing (NLP), has seen significant progress thanks to the proliferation of deep learning and the models based on Transformer \cite{vaswani2017attention}. In particular, models such as Bidirectional Auto-Regressive Transformers (BART) \cite{lewis2020bart}, Text-to-Text Transfer Transformer (T5) \cite{raffel2020exploring}, and Generative Pre-trained Transformer (GPT) \cite{radford2019language,brown2020language,achiam2023gpt}, have established themselves as key tools in this field \cite{gonzalez2022source, lam2023abstractive, kwon2023abstractive}. These models perform well in abstractive summarization, which involves generating new sentences that capture the essence of the original text rather than merely extracting and rephrasing existing sentences.

To keep track of progress in this field, many summarization datasets were proposed, such as CNN/ Daily Mail \cite{nallapati2016abstractive}, Extreme Summarization (XSum) \cite{narayan2018don}, New York Times (NY Times) \cite{hermann2015teaching}, BOOKSUM \cite{kryscinski2022booksum}, CiteSum \cite{mao2022citesum}, and CCSUM \cite{jiang2024ccsum}. 
However, most of the large summarization datasets were created for the English language, with few exceptions, such as OrangeSum \cite{eddine2021barthez} for French and the Dataset for Automatic summarization of Catalan and Spanish newspaper Articles (DACSA)
\cite{soriano2022dacsa} for Spanish. This limits the development and assessment of summarization models for other languages, impeding their performance and applicability in multilingual and diverse linguistic contexts.

Thus, we provide RoLargeSum, a dataset crawled from many news websites in Romania and the Republic of Moldova, which comprises 615,679 news articles and their summaries. In addition to the summary of each news article, we also attach to each sample of RoLargeSum, where present, the corresponding headline, keywords, dialect, and other metadata such as the domain and the author of the article. RoLargeSum contributes to the growing collection of Romanian NLP resources, complementing existing datasets for tasks such as named entity recognition \cite{dumitrescu2020introducing, avram2024histnero}, speech recognition \cite{avram2022rosac}, satire detection \cite{rogoz2021saroco, echim2023adversarial}, offensive language detection \cite{hoefels2022coroseof, matei2024enhancing, nicola2024investigating}, lip reading \cite{jitaru2020lrro, muanescu2023end}, multiword expression detection \cite{savary2018parseme, avram2023multilingual}, and emotion detection \cite{ciobotaru2022red}.

We further evaluate the performance of several BART models on RoLargeSum and further improve them by using dialect-based adversarial training \cite{ganin2016domain} and extending their input context with Unlimiformer \cite{bertsch2024unlimiformer}. Our results show that, in general, this methodology increases the performance of the BART models, with the large version of
multilingual BART (mBART) \cite{liu2020multilingual} achieving the highest overall results in most experiments. We also evaluated the capabilities of several large language models (LLMs) to produce summaries for the news articles found in RoLargeSum, and the evaluation results outlined that, in general, Romanian LLMs outperform their multilingual counterparts but still achieve robust performance. Finally, we performed a human evaluation of the best-performing model on RoLargeSum and discussed the resulting findings.

The main contributions of our work can be summarized as follows:
\begin{itemize}
    \item We introduce RoLargeSum\footnote{\url{https://github.com/avramandrei/rolargesum}}, the largest dataset for Romanian summarization, which contains more than 615K news articles, together with their summaries and headlines, annotated by dialect. RoLargeSum is also the first Romanian dataset for keyword extraction.
    \item We created an in-depth analysis of RoLargeSum and compared it with the summarization datasets available in the literature.
    \item We propose strong baselines on RoLargeSum by employing various multilingual or Romanian LLMs. We also optimize several BART models and enhance their performance through dialect adversarial training and context extension with Unlimiformer.
\end{itemize}

\section{Related Work}

\subsection{Text Summarization Datasets}

Abstractive summarization is an important and challenging task in the field of NLP. It requires models with complex natural language understanding and generation abilities to perform well. Various domains and languages provide many datasets that can be used to train and evaluate machine learning models for abstractive summarization. 

The CNN/Daily Mail dataset \cite{nallapati2016abstractive} is one of the most widely used summarization datasets, supporting both abstractive and extractive summarization for the English language. It contains 286,817 training samples, 13,368 validation samples, and 11,487 test samples, extracted from news articles written by journalists at CNN between 2007 and 2015 and the Daily Mail between 2010 and 2015.

The NY Times \cite{hermann2015teaching} is another popular abstractive summarization dataset for the English language, which contains more than 1.8 million news articles published by New York Times journalists between 1987 and 2007. However, of those 1.8 million articles, only 650K have a corresponding summary written by library scientists. Also, the NY Times corpus contains more than 1.8M tags that were manually or semi-automatically annotated.

The task of generating a brief one-sentence summary using the question “What is the article about?” is known as extreme summarization \cite{pavel2024sumhis}. XSum \cite{narayan2018don} addresses this task by providing 226,711 news articles (i.e., 204,045 samples for training, 11,332 samples for validation, and 11,334 samples for testing) with a one-sentence summary in the English language, which was collected from BBC between 2010 and 2017, covering a broad spectrum of domains.

With more than 2.8 million summarized news articles in Spanish and Catalan, DASCA \cite{soriano2022dacsa} is one of the largest and highest-quality summarization datasets available in the literature. The news was collected from 28 different news sources (i.e., 21 in Spanish and 7 in Catalan) with years of publications ranging from 2010 to 2020, resulting in 6 million news articles cleaned and filtered down using two thresholds. 

To our knowledge, RoSummary \cite{niculescu2022rosummary}
is the only summary dataset available for the Romanian language, containing 42,862 news articles gathered from one Romanian website. The news articles and their corresponding summaries were cleaned using various hand-made heuristics. 

\subsection{Text Summarization Methods}

The research community has addressed the task of automating text summarization using abstractive, extractive, or hybrid methods. On the one hand, extractive approaches generate summaries by directly selecting sentences or words from input texts \cite{narayan2018ranking,zhong2020extractive,xie2022pre,zhang2023extractive}. Generally, these techniques involve a sequential binary classification task to identify the most important sentences from the documents. They employ various criteria for this purpose, including negative log-likelihood on chosen sentences or Recall Oriented Understudy for Gist Evaluation (ROUGE) \cite{lin2004rouge} rewards within reinforcement learning frameworks \cite{yao2018deep,dong2018banditsum}.

On the other hand, abstractive summarization methods generate summaries by rephrasing the sentences that are included in the documents. Thanks to recent achievements in self-supervised learning, which involves pre-training neural networks on large amounts of texts and then fine-tuning them on a downstream task \cite{devlin2019bert,touvron2023llama}, text summarization research has increasingly shifted from extractive methods to abstractive ones \cite{wang2023t5,karim2024arabic}, which usually rely mainly on the encoder-decoder framework \cite{sutskever2014sequence}. The most effective models in this category are FlanT5 \cite{chung2024scaling}, T5, and BART, which were pre-trained on large corpora such as Common Craw \cite{wenzek2020ccnet} and the Colossal Clean Crawled Corpus \cite{raffel2020exploring}.

Finally, there are hybrid approaches that combine extractive and abstractive techniques, either implemented separately or simultaneously within the models \cite{mendes2019jointly,zhang2020pegasus}.


\section{RoLargeSum Dataset}

\subsection{Dataset Construction}

We first crawled many news websites from Romania and the Republic of Moldova that made their content publicly accessible to compile the RoLargeSum dataset. We did this by gathering the main article, its summary, its headline, its keywords, and other metadata. After that, we applied several filtering techniques to remove the artifacts from the corpus (see Appendix \ref{app:clean_rules} for more details regarding the cleaning rules).

The resulting dataset contains 615,679 samples, which we split into 605,679 for training, 5,000 for validation, and 5,000 for testing. It should also be noted that not all websites we crawled simultaneously provided summaries, headlines, or keywords for their articles. Thus, only 529,800 of the 615,679 collected samples had summaries, 613,836 headlines, and 426,455 keywords.  However, we ensured that the samples in the validation and testing sets contained these fields.

\subsection{Dataset Statistics}

\begin{table*}[t]
\resizebox{\textwidth}{!}{
\begin{tabular}{l|c|cc|cc|cc}
     \toprule
     \multirow{ 2}{*}{\textbf{Dataset}} & \multirow{ 2}{*}{\textbf{Train/Val/Test}} & \multicolumn{2}{c|}{\textbf{Avg. News  Len.}} & \multicolumn{2}{c|}{\textbf{Avg. Sum.  Len.}} & \multicolumn{2}{c}{\textbf{Vocab. Size}} \\
     & & \textbf{Words} & \textbf{Sents.} & \textbf{Words} & \textbf{Sents.} & \textbf{News} & \textbf{Sums.} \\
     \midrule
     CNN & 90.3K/1.2K/1.1K & 760.50 & 33.98 & 45.70 & 3.58 & 34K & 89K \\
     DailyMail & 197K/12.1K/10.4K & 653.33 & 29.33 & 54.65 & 3.86 & 564K & 180K \\
     NY Times & 590K/32.7K/32.7K & 800.04 & 35.55 & 45.54 & 2.44 & 1.2M & 293K \\
     XSum & 204K/11.3K/11.3K & 431.07 & 19.77 & 23.26 & 1.00 & 399K & 81K \\
     DACSA-Spanish & 1.8M/104K/104K & 644.64 & 23.44 & 28.45 & 1.24 & 3.2M & 516K \\
     DACSA-Catalan & 636K/35.3K/35.3K& 507.74 & 17.71 & 24.09 & 1.17 & 1.3M & 223K \\
     OrangeSum-Summary & 21.4K/1.5K/1.5K & 350.01 &  12.06 & 32.12 & 1.43 & 420K & 71K \\
     OrangeSum-Headline & 30.6K/1.5K/1.5K & 315.31 & 10.87 & 11.42 & 1.00 & 483K & 43K \\
     RoSummary-Summary & 38.7K/2.0K/2.1K & 329.30 & 8.39 & 56.70 & 3.26 & 329K & 98K \\
     RoSummary-Headline & 38.7K/2.0K/2.1K & 329.30 & 8.39 & 18.44 & 1.53 & 329K & 49K \\
     \midrule
     RoLargeSum-Summary & 519K/5.0K/5.0K & 336.88 & 9.96 & 46.13 & 1.57 & 1.3M & 295K \\
     RoLargeSum-Headline & 603K/5.0K/5.0K & 337.11 & 9.82 & 15.87 & 1.24 & 1.4M & 153K \\
     RoLargeSum-Keywords & 416K/5.0K/5.0K & 337.30 & 9.31 & 8.08 & 1.01 & 1.2M & 83K \\
     \bottomrule
\end{tabular}
}
\centering
\caption{Comparison of RoLargeSum with other summarization datasets available in the literature. We outline the train/validation/test distribution of the data, the average numbers of words and sentences in the news articles and their corresponding summaries, and the vocabulary size for both the news articles and their summaries. Part of the statistics were taken from \cite{eddine2021barthez}.}
\label{tab:comp_stats}
\end{table*}

\begin{figure}
    \centering
    \includegraphics[width=0.49\textwidth]{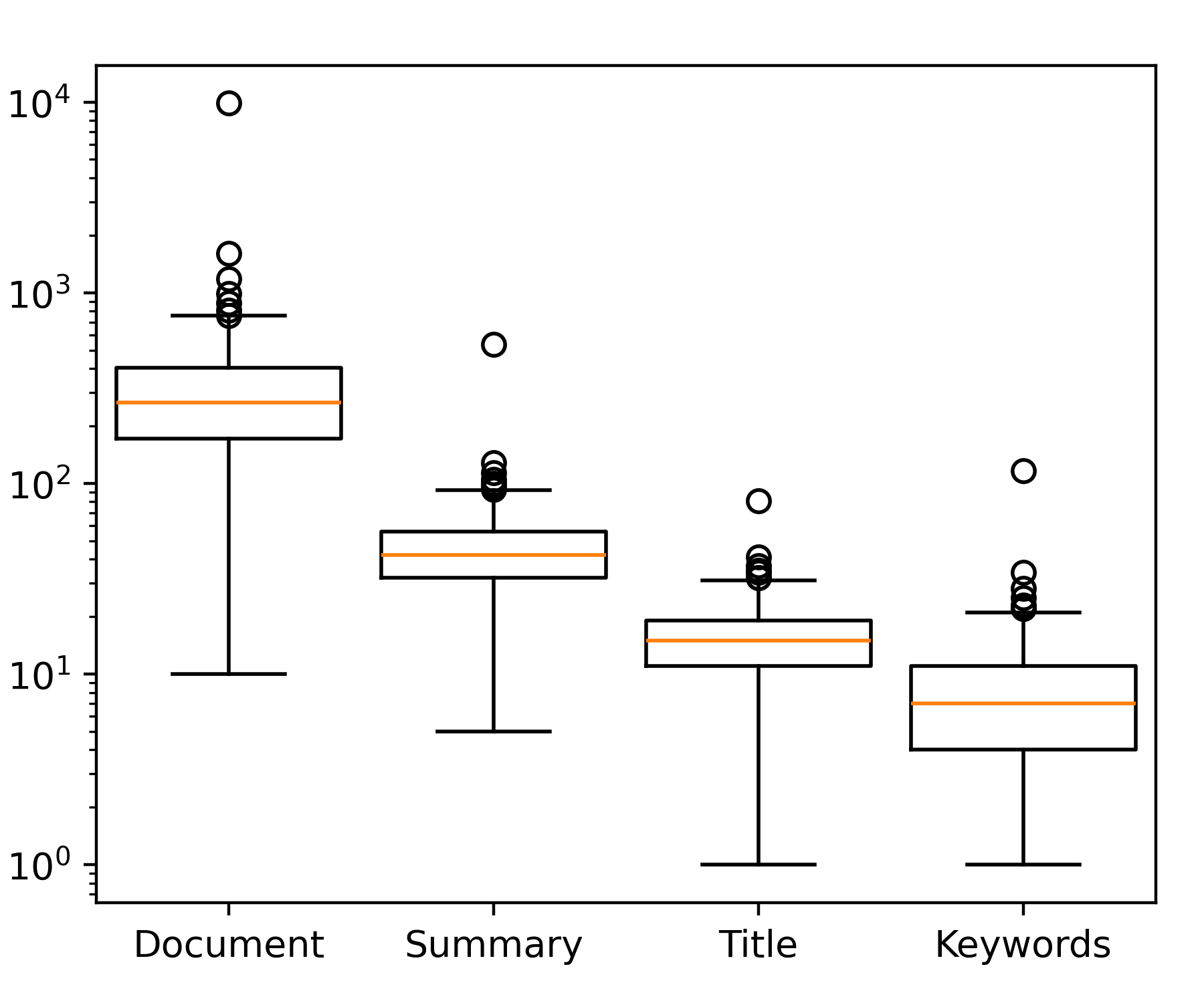}
    \caption{Boxplots depicting the numbers of words for the documents, summaries, headlines, and keywords in the RoLargeSum dataset.}
    \label{fig:tok_stats}
\end{figure}

Figure \ref{fig:tok_stats} shows the RoLargeSum token length statistics. The average length of the gathered news articles is 337 tokens, with a minimum of 10 tokens and a maximum of 9,851 tokens; their summaries are 46 tokens long, with a minimum of 5 tokens and a maximum of 537 tokens; their headlines are 15 tokens long, with a minimum of 1 token and a maximum of 81 tokens; and lastly, each news article has an average of 8 keywords, with a minimum of 1 keyword and a maximum of 116 keywords.

We further classified the dataset by its dialect using the top-level domain (``.ro'' - corresponding to newspapers from Romania or ``.md'' - corresponding to newspapers from the Republic of Moldova) of the websites from which the samples were collected. With 464,465 samples belonging to the Moldavian news articles and 151,214 samples belonging to the Romanian news articles, the dataset's dialect distribution is highly imbalanced by a ratio of roughly 3:1. We also outline a dialect-based term frequency-inverse document frequency (TF-IDF) \cite{salton1988term} analysis in Appendix \ref{app:tfidf}.

\subsection{Dataset Comparison}

We further compare RoLargeSum with several popular summarization datasets available in the literature for English (i.e., CNN, DailyMail, NY Times, and XSum), Spanish (i.e., DACSA), Catalan (i.e., DACSA), French (i.e., OrangeSum), as well as Romanian (i.e., RoSummary). The results of the statistical analysis are shown in Table \ref{tab:comp_stats} (see Appendix \ref{app:dialect_stats} for the statistics based on Romanian and Moldavian news articles). As can be observed, with 529K documents, RoLargeSum is the third largest dataset in this comparison, falling behind NY Times with 655K samples and DACSA with 2M samples for Spanish and 706K for Catalan. 
 
On the other hand, the average number of words found in RoLargeSum's documents is on the lower end, with approximately 337 words per document, surpassing only the headline generation subset of OrangeSum and the RoSummary dataset by a little margin. The average number of words in the summaries is the third highest, behind only DailyMail and RoSummary. Finally, with 1.3M words in the vocabulary, RoLargeSum has a vocabulary size comparable to other datasets of similar size.

\subsection{Proposed Subtasks}

We propose the following subtasks on RoLargeSum:
\begin{itemize}
  \item RO+MD - Romanian and Moldavian summarization, as well as headline and keyword generation. The subtask generates the summary, headline, and keywords for the whole corpus, ignoring the dialect label.
  
  \item RO - Romanian intra-dialect summarization, as well as headline and keyword generation. The subtask generates the summary, headline, and keywords only for the news articles from Romania.
  
  \item MD - Moldavian intra-dialect summarization, as well as headline and keyword generation. The subtask generates the summary, headline,
  and keywords only for Moldavian news samples.
  
  \item RO→MD - cross-dialect summarization, together with headline and keyword generation. The subtask generates the summary, headline, and keywords for Moldavian news samples using a model trained on news articles from Romania. 
  
  \item MD→RO - cross-dialect summarization, together with headline and keyword generation. The subtask generates the summary, headline, and keywords for news articles from Romania using a model trained on Moldavian news samples. 
\end{itemize}



\section{Baseline Models}

\subsection{Fine-tuned Models}

\begin{figure}[!ht]
    \centering
    \includegraphics[width=0.47\textwidth]{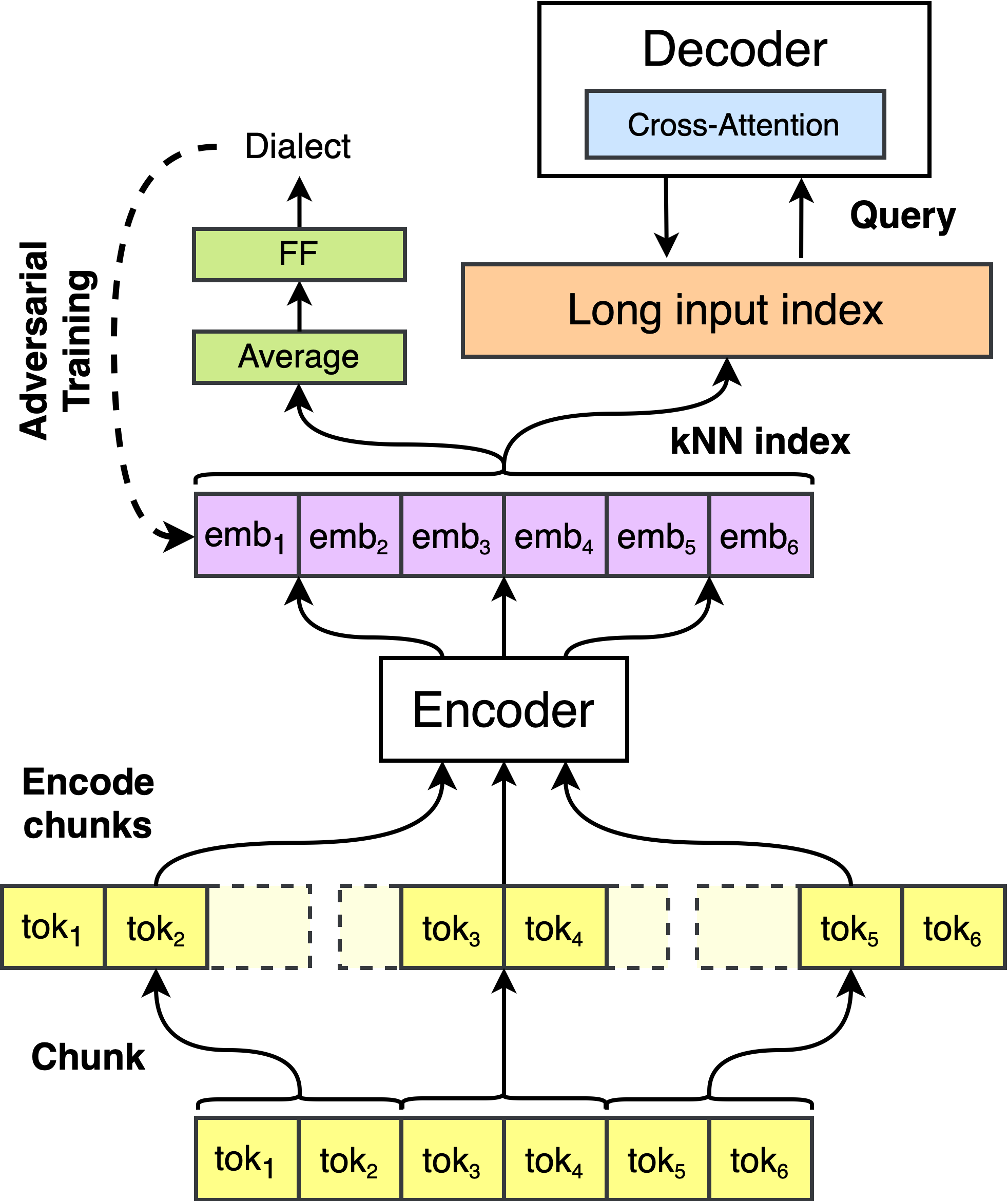}
    \caption{The proposed baseline architecture used to generate summaries, headlines, and keywords in RoLargeSum. We chunk a set of tokens and add them to each chunk in the neighbouring context.  
    After receiving the chunks, the encoder generates the corresponding embeddings, which are then used to build a kNN index for the cross-attention mechanism of the decoder. Finally, we also adversarially train a feed-forward network to detect the dialect of the input text, which helps the encoder produce dialect-independent embeddings.}
    \label{fig:syst_arch}
\end{figure}

Figure \ref{fig:syst_arch} shows our overall architecture, including an encoder-decoder system wrapped with Unlimiformer and adversarially trained to produce dialect-independent embeddings with the encoder. The architecture is used to fine-tune the BART variations in our experiments. 

\paragraph{BART} We establish several strong baselines on RoLargeSum by fine-tuning and evaluating the performance to generate summaries, headlines, and keywords for the base and large versions of BART (i.e., BART-base and BART-large, respectively) and the large multilingual version of BART (i.e., mBART-large) \cite{liu2020multilingual} on each of the dialect-based subtasks. An essential aspect of the fine-tuning process is that both BART-base and BART-large have been trained only in English text, with little to no Romanian. As recommended by \citet{lewis2020bart} for machine translation, we develop a new Romanian vocabulary and randomly initialize the embedding layer included in both encoders and decoders to avoid this problem.

\paragraph{Unlimiformer} Unlimiformer \cite{bertsch2024unlimiformer} is a versatile method that can be applied to any pre-trained encoder-decoder Transformer such as BART and T5. It does this by redesigning the cross-attention mechanism to a single k-Nearest Neighbor (kNN) index, using the attention dot-product scores, which are the distances provided by the kNN algorithm \cite{bertsch2024unlimiformer}. We emphasize the ability of this algorithm to index inputs of unlimited length since each attention head in each decoder layer only retrieves the top k keys rather than all of the encoder's keys \cite{bertsch2024unlimiformer}.


We use Unlimiformer since, as depicted in Figure \ref{fig:tok_stats}, the documents can include up to 10k tokens, which is significantly longer than the context length of BART. Thus, we perform two experiments for each BART variant, namely
1) we truncate all the documents in RoLargeSum that have more tokens than the supported maximum context length and fine-tune each BART variant without any modifications to the model, and 2) we truncate the documents during training and inject the Unlimiformer wrapper into the BART model during inference to process the full-length input sequences.

\paragraph{Adversarial Training} We further adversarially train \cite{ganin2016domain} the BART variants to produce dialect-independent features in the encoder, which we employ only for the RO+MD subtask. To do that, we take the average of the embeddings produced by the encoder and add a feed-forward layer on top of the resulting vector, which is responsible for identifying whether the input news article is from Romania or the Republic of Moldova.
Then, during backpropagation, we either 1) reverse the gradients before backpropagating them into the encoder using a gradient reversal layer \cite{ganin2016domain}, or 2) directly change the sign of the loss function \cite{avram2024histnero}.

More specifically, we take the average of the embeddings $e_i$ produced by the encoder for the input sequence and employ a linear layer with weights $\theta_D$ and bias $b_D$ to get the dialect label $\hat{y}$, as formulated in Equation \ref{eq:dialect_ff}.:

\begin{equation}
    \hat{y} = \sigma(\theta_D \bar{e} + b_D)
    \label{eq:dialect_ff}
\end{equation}
where $\sigma$ is the sigmoid activation function and $\bar{e}$ is the average of the embeddings $e_i$. Then, we employ the binary cross-entropy loss as the objective function $\mathcal{L}_d$, which can be formulated as:

\begin{equation}
    \mathcal{L}_D = - \sum_{i=1}^{|\mathcal{D}|} (y_i \log(\hat{y}_i) + (1 - y_i)\log(1 - \hat{y}_i))
\end{equation}
where $|\mathcal{D}|$ is the number of documents in the dataset, and $y_i$ is equal to 1 if the news article is from Romania and 0 otherwise.

We apply the adversarial training to this system by reversing the feed-forward gradient as follows:

\begin{equation}
    \theta_D = \theta_D - \alpha \frac{\partial \mathcal{L}_D}{\partial \theta_D}
\end{equation}

\begin{equation}
    \theta_{dec} = \theta_{dec} - \alpha \frac{\partial \mathcal{L}_{dec}}{\partial \theta_{dec}}
\end{equation}

\begin{equation}
    \theta_{enc} = \theta_{enc} - \alpha \Bigg(\frac{\partial \mathcal{L}_{dec}}{\partial \theta_{dec}} - \lambda_{GR} \frac{\partial \mathcal{L}_{D}}{\partial \theta_{D}}\Bigg)
\end{equation}
where $\alpha$ is the learning rate, $\theta_{dec}$ are the parameters of the decoder, $\mathcal{L}_{dec}$ is the loss function used by the decoder, $\theta_{enc}$ are the parameters of the encoder, and $\lambda_{GR}$ is a scaling hyperparameter.

For the loss reversal adversarial training, we simply change the sign of the loss $\mathcal{L}_D$ and multiply it by another scaling hyperparameter $\lambda_{LR}$.

\subsection{Large Langauge Models}

We also evaluate the performance of various Romanian and multilingual LLMs from the Llama, Mistral, and Gemma families. We provide more details in Appendix \ref{app:llms}. In addition, we provide more information on the prompt we used for summary generation in Appendix \ref{app:prompt}. 

\section{Evaluation}

We measure the performance of all models using the ROUGE score \cite{lin2003automatic} for which we provide more information in Appendix \ref{app:rouge}. In addition, more details on implementing our methodology can be found in Appendix \ref{app:implem}.

\subsection{Results for BART Fine-Tuning}

\begin{table*}[!ht]
  \centering
  \begin{tabular}{l|ccc|ccc|ccc}
    \toprule
    \multirow{ 2}{*}{\textbf{Model}} & \multicolumn{3}{c|}{\textbf{Summary}} & \multicolumn{3}{c|}{\textbf{Headline}} & \multicolumn{3}{c}{\textbf{Keywords}} \\
    & \textbf{R-1} & \textbf{R-2} & \textbf{R-L} & \textbf{R-1} & \textbf{R-2} & \textbf{R-L} & \textbf{R-1} & \textbf{R-2} & \textbf{R-L} \\
    \midrule
    \multicolumn{10}{c}{\textbf{\textit{RO+MD}}} \\
    \midrule
    BART-base & 37.56 & 18.62 & 28.68 & 35.32 & 18.22 & 30.40 & 51.27 & 15.10 & 48.86 \\
     \hspace{0.7ex} /w Unlimiformer & 37.15 & 18.95 & 28.59 & 34.97 & 19.20 & 29.26 & 52.65 & 15.29 & 48.97 \\
     \hspace{2.5ex} /w Grad. Rev. & 38.59 & 19.06 & 18.90 & 35.25 & 18.84 & 30.11 & 52.79 & 15.40 & 49.08 \\
     \hspace{2.5ex} /w Loss Rev. & 38.63 & 19.07 & 18.89 & 35.33 & 19.17 & 30.34 & 53.00 & 15.72 & 49.21 \\
     BART-large & 39.18 & 19.92 & 29.92 & 37.08 & 19.58 & 31.65 & 56.28 & 29.40 & 54.79 \\
     \hspace{0.7ex} /w Unlimiformer & 39.19 & 18.61 & 29.95 & 37.73 & 21.99 & 33.75 & 55.41 & 29.10 & 54.18 \\
     \hspace{2.5ex} /w Grad. Rev. & 39.35 & 19.79 & 30.07 & 37.76 & 22.04 & 33.80 & 55.40 & 29.05 & 54.19 \\
     \hspace{2.5ex} /w Loss Rev. & 39.24 & 19.67 & 30.19 & 38.01 & 22.16 & 33.92 & 55.46 & 29.09 & 54.15 \\
    mBART-large & 41.42 & 21.65 & 32.27 & 39.89 & 20.37 & 35.30 & 57.97 & 33.65 & 55.86 \\
     \hspace{0.7ex} /w Unlimiformer & 44.38 & 23.86 & 35.91 & 43.09 & 23.62 & 41.37 & 59.76 & 33.95 & 56.18 \\
     \hspace{2.5ex} /w Grad. Rev. & 44.49 & 23.98 & 36.02 & 43.25 & \textbf{23.70} & \textbf{41.47} & 59.72 & 33.90 & 56.15 \\
     \hspace{2.5ex} /w Loss Rev. & \textbf{44.57} & \textbf{24.03} & \textbf{36.10} & \textbf{43.31} & 23.64 & 41.12 & \textbf{59.80} & \textbf{34.0}1 & \textbf{56.29} \\
     \midrule
     \multicolumn{10}{c}{\textbf{\textit{RO}}} \\
    \midrule
     BART-base & 36.28 & 17.58 & 27.68 & 33.62 & 16.59 & 28.50 & 41.42 & 17.73 & 38.36 \\
     \hspace{0.7ex} /w Unlimiformer & 37.71 & 18.11 & 28.43 & 35.18 & 17.40 & 29.85 & 41.32 & 17.59 & 38.33 \\
     BART-large & 38.14 & 18.57 & 28.42 & 36.28 & 18.61 & 30.50 & 42.54 & 18.33 & 38.73 \\
     \hspace{0.7ex} /w Unlimiformer & 39.19 & 18.21 & 29.95 & 36.73 & \textbf{18.99} & 32.75 & 42.18 & 18.06 & 37.98 \\
     mBART-large & 40.34 & 20.58 & 31.31 & 38.49 & 18.84 & \textbf{33.70} & 50.52 & \textbf{21.67} & 47.42 \\
     \hspace{0.7ex} /w Unlimiformer & \textbf{41.38} & \textbf{20.88} & \textbf{31.91} & \textbf{39.09} & 18.62 & 32.37 & \textbf{51.12} & 21.56 & \textbf{47.83} \\
     \midrule
     \multicolumn{10}{c}{\textbf{\textit{MD}}} \\
    \midrule
     BART-base & 36.75 & 18.13 & 28.57 & 33.08 & 16.64 & 28.75 & 63.56 & 17.48 & 63.32 \\
     \hspace{0.7ex} /w Unlimiformer & 36.22 & 17.51 & 27.81 & 33.37 & 16.71 & 29.07 & 63.86 & 18.46 & 63.49 \\
     BART-large & 38.46 & 19.38 & 29.60 & 35.59 & 17.75 & 29.68 & 63.88 & 17.59 & 63.64 \\
     \hspace{0.7ex} /w Unlimiformer & 39.09 & 19.52 & 29.82 & 36.08 & 19.36 & 31.28 & 64.62 & 18.02 & 64.36 \\
     mBART-large & 42.00 & 22.52 & 33.11 & 38.94 & 19.49 & 34.49 & \textbf{72.97} & \textbf{20.84} & \textbf{72.76} \\
     \hspace{0.7ex} /w Unlimiformer & \textbf{42.48} & \textbf{22.85} & \textbf{33.61} & \textbf{39.35} & \textbf{19.86} & \textbf{35.02} & 72.35 & 20.59 & 72.28 \\
     \midrule
     \multicolumn{10}{c}{\textbf{\textit{RO→MD}}} \\
    \midrule
    BART-base & 28.56 & 9.89 & 20.19 & 25.94 & 10.17 & 21.07 & 37.84 & 13.49 & 35.06 \\
     \hspace{0.7ex} /w Unlimiformer & 29.25 & 10.76 & 20.77 & 26.86 & 10.96 & 21.91 & 37.47 & 13.25 & 34.89 \\
     BART-large & 30.99 & 11.05 & 22.18 & 27.77 & 11.16 & 22.37 & 38.11 & 14.03 & 36.57 \\
     \hspace{0.7ex} /w Unlimiformer & 31.40 & 11.69 & 21.78 & 27.86 & 11.20 & 22.45 & 38.37 & 14.30 & 36.79 \\
     mBART-large & \textbf{32.93} & \textbf{12.26} & 23.98 & \textbf{31.92} & 12.11 & 27.58 & 40.20 & 16.07 & \textbf{40.72} \\
     \hspace{0.7ex} /w Unlimiformer & 32.87 & 12.14 & \textbf{24.73} & 31.35 & \textbf{12.43} & \textbf{27.88} & \textbf{40.34} & \textbf{16.10} & 40.70 \\
     \midrule
     \multicolumn{10}{c}{\textbf{\textit{MD→RO}}} \\
    \midrule
     BART-base & 33.70 & 14.91 & 24.85 & 31.47 & 14.80 & 26.89 & 23.22 & 7.12 & 20.82 \\
     \hspace{0.7ex} /w Unlimiformer & 33.07 & 14.50 & 21.17 & 31.16 & 14.69 & 25.95 & 23.32 & 7.46 & 21.08 \\
     BART-large & 35.46 & 16.37 & 26.26 & 34.57 & 17.23 & 29.33 & 24.38 & 7.46 & 21.88 \\
     \hspace{0.7ex} /w Unlimiformer & 36.35 & 17.58 & 27.14 & 35.23 & 18.19 & 30.32 & 23.68 & 7.04 & 21.16 \\
     mBART-large & 37.19 & \textbf{17.11} & \textbf{28.38} & 36.76 & 16.88 & \textbf{32.09} & 28.10 & 8.81 & \textbf{27.20} \\
     \hspace{0.7ex} /w Unlimiformer & \textbf{38.13} & 16.86 & 28.28 & \textbf{37.16} & \textbf{16.97} & 31.86 & \textbf{28.21} & \textbf{8.87} & 27.08 \\
     \bottomrule
  \end{tabular}
  \caption{Evaluation results of the BART models for summary, headline, and keyword generation for the four subtasks proposed on RoLargeSum.}
  \label{tab:bart_results}
\end{table*}

The results of the BART models on all four subtasks proposed for RoLargeSum are depicted in Table \ref{tab:bart_results}. In general, the mBART-large model obtained the greatest scores.  We believe that this is due to the pre-training data, this model being trained specifically on Romanian, among other languages, which naturally boosted its understanding of the input text, allowing it to outperform the other two evaluated BART models.

\paragraph{RO+MD} The highest ROUGE scores for the RO+MD subtask in summarization were obtained by the mBART model that employed both the Unlimiformer wrapper and the loss reversal adversarial training method. It obtained an R-1 of 44.57, an R-2 of 24.03, and an R-L of 36.10. This methodology also obtained the highest scores for keyword generation, with an R-1 of 59.80, an R-2 of 34.01, and an R-L of 56.29. Still, it obtained only the highest R-1 for headline generation with 43.31, being outperformed by the mBART with the Unlimiformer and the gradient reversal algorithms on the R-2 and R-L scores.

\paragraph{RO} The best scores for summary generation on the subset that used only the news articles from Romania were obtained by the mBART-large model with the Unlimiformer wrapper, which obtained an R-1 of 41.38, an R-2 of 20.88 and an R-L of 31.91. Alternatively, this methodology obtained the highest R-1 score for headline generation, with 39.09. However, it fell behind the BART-large variant with Unlimiformer in R-2, which obtained a score of 18.99, and in R-L compared to the plain variant of mBART-large, which achieved a score of 33.70. Finally, mBART-large also obtained the highest score on keyword generation by employing the Unlimiformer wrapper on R-1 with 51.12, R-L with 47.83, and without it on R-2 with 21.67.

\paragraph{MD} The highest ROUGE scores for summarization on the RoLargeSum subset that contained only the news articles from the Republic of Moldova were also obtained by the mBART-large model with the Unlimiformer wrapper, obtaining an R-1 score of 42.48, an R-2 score of 22.85, and an R-L score of 33.61. This methodology also obtained the best score for headline generation, with an R-1 of 39.35, an R-2 of 19.86, and an R-L of 35.02. However, it was outperformed by the plain mBART-large in keyword generation on all three metrics, which achieved an R-1 of 72.97, an R-2 of 20.84, and an R-L of 72.76.

\paragraph{RO→MD} The results are somewhat mixed on the subtask RO→MD, with the highest ROUGE scores oscillating between mBART-large with and without Unlimiformer. Thus, for summarization, the plain mBART-large obtained the highest scores R-1 and R-2 with 32.93 and 12.26, respectively, and by wrapping the model with Unlimiformer, its results increased to 24.73. For headline generation, the highest R-1 score was obtained by the plain mBART-large, and the highest R-2 and R-L scores were achieved by the mBART-large with Unlimiformer (i.e., 12.43 and 27.88, respectively). Finally, the Unlimiformer boosted the performance of mBART-large in R-1 with 40.34 and in R-2 with 16.10, but not in R-L, where the plain mBART-large performed better, obtaining a score of 40.72.

\paragraph{MD→RO}  The results are again mixed for this subtask, similar to the RO→MD subtask. mBART-large obtained the top score R-1 (i.e., 38.13) for summarization by employing Unlimiformer, while the plain version obtained the highest scores R-2 and R-L. For headline generation, the Unlimiformer wrapper helped improve the results for the R-1 and R-2 scores, with mBART-large obtaining 37.16 and 16.97, respectively, but it did not help on the R-L score where the plain version achieved an R-L score of 32.09. The same pattern also applies to keyword generation, namely the highest R-1 and R-2 scores were obtained by mBART-large with Unlimiformer, and the highest R-L scores were obtained without this wrapper.

\subsection{Results for LLMs}

\begin{table}[]
    \centering
    \begin{tabular}{l|ccc}
         \toprule
    \textbf{Model} & \textbf{R-1} & \textbf{R-2} & \textbf{R-L} \\
    \midrule
    Llama-2-7B & 22.23 & 10.04 & 16.83 \\
    Llama-2-7B-It & \textbf{30.01} & \textbf{13.39} & \textbf{20.93} \\
    Llama-3-8B & 25.81 & 12.23 & 18.79 \\
    Llama-3-8B-It & 16.84 & 7.55 & 11.79 \\
    Llama-3.1-8B & 27.07 & 12.71 & 19.39 \\
    Llama-3.1-8B-It & 28.72 & 13.16 & 20.00 \\
    \midrule
    Mistral-7B-v0.1 & 7.45 & 4.14 & 6.09 \\
    Mistral-7B-It-v0.1 & 27.27 & 12.54 & 19.16 \\
    Mistral-7B-v0.2 & 9.61 & 5.14 & 7.78 \\
    Mistral-7B-It-v0.2 & 27.97 & 11.73 & 18.66 \\
    Mistral-7B-v0.3 & 9.63 & 5.13 & 7.82 \\
    Mistral-7B-It-v0.3 & \textbf{31.16} & \textbf{14.12} & \textbf{21.37} \\
    \midrule
    Gemma-7B & 19.12 & 09.85 & 14.49 \\
    Gemma-7B-It & 03.60 & 01.41 & 02.85 \\
    Gemma-1.1-7B-It & 04.88 & 02.17 & 04.09 \\
    Gemma2-9B & \textbf{24.74} & \textbf{13.20} & \textbf{19.12} \\
    Gemma2-9B-It & 05.08 & 03.55 & 04.76 \\
    \midrule
    RoLlama2-7B & 29.00 & \textbf{15.25} & \textbf{22.23} \\
    RoLlama2-7B-It & \textbf{30.14} & 13.88 & 21.40 \\
    RoQLlama2-7B & 25.34 & 11.37 & 18.88 \\
    RoLlama3-8B-It & 29.39 & 13.01 & 20.10 \\
    RoMistral-7B-It & 29.77 & 13.68 & 20.63 \\
    RoGemma-7B-It & 26.66 & 12.14 & 18.61 \\
    \bottomrule
    \end{tabular}
    \caption{Evaluation results of multilingual and Romanian LLMs from the Llama, Mistral, and Gemma families.}
    \label{tab:llms_results}
\end{table}

The results of the evaluation of LLMs for the RO+MD summarization task on RoLargeSum are presented in Table \ref{tab:llms_results}. The highest R-1 score was obtained by the Mistral-7B-It-v0.3 model with 31.16, while the highest R-2 and R-L scores were obtained by RoLlama2-7B with 15.25 and 22.23, respectively, underperforming compared to all fine-tuned BART models in this task. 

Additionally, we can see that in four of five instances, the Romanian variants of the model families performed better than their multilingual counterparts on all metrics: (1) RoGemma-7B-It outperformed all Gemma models, (2) RoLlama3-8B-it outperformed all Llama-3 models, (3) RoLlama2-7B and RoQLlama2-7B outperformed Llama-2-7B, (4) RoLlama2-7B-it outperformed Llama-2-7B-It, with RoMistral-7B-It being surpassed only by Mistral-7B-It-v0.3.

\subsection{Human Evaluation}

\begin{table}
    \centering
    \resizebox{0.475\textwidth}{!}{
    \begin{tabular}{l|ccc}
         \toprule
    \textbf{Metric} & \textbf{Summary} & \textbf{Headline} & \textbf{Keywords} \\
    \midrule
    Coherence & 4.3 & 4.8 & - \\
    Consistency & 3.9 & 4.1 & -\\
    Coverage & 3.4 & 4.3 & 4.4 \\
    Fluency & 4.4 & 4.5 & - \\
    Overall & 3.9 & 4.3 & 4.3 \\
    \bottomrule
    \end{tabular}
    }
    \caption{Human evaluation of the best-performing model on RoLargeSum.}
    \label{tab:hum_annotation}
\end{table}

We further manually analyze the ability of the best-performing system, mBART-large + Unlimiformer + Loss Reversal, to generate summaries, headlines, and keywords on a subset of 100 documents extracted from the RoLargeSum's test set by following \cite{el2024arabic,barta2024news}. We accomplish this using five annotators who are required to assign a score ranging from 1 to 5 in the following categories to each generated summary, headline, and keywords (SHKs)\footnote{We do not compute the coherence, consistency, and fluency scores for keywords since these categories are not representative for an enumeration of words.}:
\begin{itemize}
    \item \textit{Coherence}: the generated SHKs are well-organized and easy to read.
    \item \textit{Consistency}: the generated SHKs are consistent with the information found in the news article.
    \item \textit{Coverage}: the main ideas of the news article are covered by the generated SHKs.
    \item \textit{Fluency}: the generated SHKs are written in good Romanian.
    \item \textit{Overall}: how good the SHKs are overall.
\end{itemize}

The results are depicted in Table \ref{tab:hum_annotation}. We can observe that both the generated summaries and headlines are somewhat coherent and use the Romanian language correctly. However, they are less consistent with their respective news articles, and the summaries usually have low coverage of the full news articles. At the same time, keywords and headlines typically cover the full content. Finally, the headlines and keywords obtained a relatively high overall score, while, in general, the generated summaries were considered by the annotators less representative of the news articles.

\section{Conclusion}

This paper introduces RoLargeSum, a large dataset with more than 615K samples that can generate summaries, headlines, and keywords in the Romanian language. The data was retrieved from multiple news websites and cleaned to ensure high quality. We further distinguished between news articles written by authors in the Republic of Moldova and Romania, which allowed us to boost the performance of the evaluated BART models, and, together with the Unlimiformer wrapper, we managed to propose several strong baselines that other researchers can use to either build upon or compare to. Finally, we intend to incorporate both the RoLargeSum dataset and the results obtained by the evaluated models into the LiRo benchmark \cite{dumitrescu_nips}.

\section{Limitations}

The distribution of dialects in RoLargeSum is not uniform. This difference is mainly due to the varying number of newspapers in Romania and the Republic of Moldova. This unbalanced distribution highlights the importance of considering geographic and media-related factors when conducting data-driven research or analysis on this dataset.

\section{Ethical Considerations}

To align with ethical research practices, it is essential to clarify that the data collection process for the RoLargeSum dataset fully respects copyrights and intellectual property rights. The dataset consists of articles and their metadata from publicly accessible websites of major Romanian and Moldavian newspapers. Each news article in the dataset is adequately cited, including its source, with an associated URL for verification and direct reference. The data is used exclusively for academic and research purposes to advance the field of Romanian text summarization. The extraction and use of these data comply with all relevant ethical guidelines, ensuring the journalistic integrity of the original articles and their authors is preserved. Thus, the RoLargeSum dataset is intended to serve as a high-quality and diverse resource for research while upholding all ethical and legal standards.

\section*{Acknowledgements}
This work was supported by the National University of Science and Technology POLITEHNICA Bucharest through the PubArt program and a grant from the National Program for Research of the National Association of Technical Universities, GNAC ARUT 2023.

\bibliography{custom}

\begin{thebibliography}{62}
\providecommand{\natexlab}[1]{#1}

\bibitem[{Achiam et~al.(2023)Achiam, Adler, Agarwal, Ahmad, Akkaya, Aleman, Almeida, Altenschmidt, Altman, Anadkat et~al.}]{achiam2023gpt}
Josh Achiam, Steven Adler, Sandhini Agarwal, Lama Ahmad, Ilge Akkaya, Florencia~Leoni Aleman, Diogo Almeida, Janko Altenschmidt, Sam Altman, Shyamal Anadkat, et~al. 2023.
\newblock Gpt-4 technical report.
\newblock \emph{arXiv preprint arXiv:2303.08774}.

\bibitem[{Avram et~al.(2024)Avram, Iuga, Manolache, Matei, Micliu{\c{s}}, Muntean, Sorlescu, {\c{S}}erban, Urse, P{\u{a}}i{\c{s}} et~al.}]{avram2024histnero}
Andrei-Marius Avram, Andreea Iuga, George-Vlad Manolache, Vlad-Cristian Matei, R{\u{a}}zvan-Gabriel Micliu{\c{s}}, Vlad-Andrei Muntean, Manuel-Petru Sorlescu, Drago-Andrei {\c{S}}erban, Adrian-Dinu Urse, Vasile P{\u{a}}i{\c{s}}, et~al. 2024.
\newblock Histnero: Historical named entity recognition for the romanian language.
\newblock In \emph{International Conference on Document Analysis and Recognition}, pages 126--144. Springer.

\bibitem[{Avram et~al.(2023)Avram, Mititelu, P{\u{a}}iș, Cercel, and Tr{\u{a}}ușan-Matu}]{avram2023multilingual}
Andrei-Marius Avram, Verginica~Barbu Mititelu, Vasile P{\u{a}}iș, Dumitru-Clementin Cercel, and Ștefan Tr{\u{a}}ușan-Matu. 2023.
\newblock Multilingual multiword expression identification using lateral inhibition and domain adaptation.
\newblock \emph{Mathematics}, 11(11):2548.

\bibitem[{Avram et~al.(2022)Avram, Nichita, Bartusica, and Mihai}]{avram2022rosac}
Andrei-Marius Avram, Mihai-Virgil Nichita, Razvan-George Bartusica, and M{\u{a}}d{\u{a}}lin-Virgil Mihai. 2022.
\newblock Rosac: A speech corpus for transcribing romanian emergency calls.
\newblock In \emph{2022 14th International Conference on Communications (COMM)}, pages 1--5. IEEE.

\bibitem[{Barta et~al.(2024)Barta, Lakatos, Nagy, Nyist, and {\'A}cs}]{barta2024news}
Botond Barta, Dorina Lakatos, Attila Nagy, Mil{\'a}n~Konor Nyist, and Judit {\'A}cs. 2024.
\newblock From news to summaries: Building a hungarian corpus for extractive and abstractive summarization.
\newblock In \emph{Proceedings of the 2024 Joint International Conference on Computational Linguistics, Language Resources and Evaluation (LREC-COLING 2024)}, pages 7503--7509.

\bibitem[{Bertsch et~al.(2024)Bertsch, Alon, Neubig, and Gormley}]{bertsch2024unlimiformer}
Amanda Bertsch, Uri Alon, Graham Neubig, and Matthew Gormley. 2024.
\newblock Unlimiformer: Long-range transformers with unlimited length input.
\newblock \emph{Advances in Neural Information Processing Systems}, 36.

\bibitem[{Brown et~al.(2020)Brown, Mann, Ryder, Subbiah, Kaplan, Dhariwal, Neelakantan, Shyam, Sastry, Askell et~al.}]{brown2020language}
Tom Brown, Benjamin Mann, Nick Ryder, Melanie Subbiah, Jared~D Kaplan, Prafulla Dhariwal, Arvind Neelakantan, Pranav Shyam, Girish Sastry, Amanda Askell, et~al. 2020.
\newblock Language models are few-shot learners.
\newblock \emph{Advances in neural information processing systems}, 33:1877--1901.

\bibitem[{Chung et~al.(2024)Chung, Hou, Longpre, Zoph, Tay, Fedus, Li, Wang, Dehghani, Brahma et~al.}]{chung2024scaling}
Hyung~Won Chung, Le~Hou, Shayne Longpre, Barret Zoph, Yi~Tay, William Fedus, Yunxuan Li, Xuezhi Wang, Mostafa Dehghani, Siddhartha Brahma, et~al. 2024.
\newblock Scaling instruction-finetuned language models.
\newblock \emph{Journal of Machine Learning Research}, 25(70):1--53.

\bibitem[{Ciobotaru et~al.(2022)Ciobotaru, Constantinescu, Dinu, and Dumitrescu}]{ciobotaru2022red}
Alexandra Ciobotaru, Mihai~Vlad Constantinescu, Liviu~P Dinu, and Stefan Dumitrescu. 2022.
\newblock Red v2: enhancing red dataset for multi-label emotion detection.
\newblock In \emph{Proceedings of the Thirteenth Language Resources and Evaluation Conference}, pages 1392--1399.

\bibitem[{Devlin et~al.(2019)Devlin, Chang, Lee, and Toutanova}]{devlin2019bert}
Jacob Devlin, Ming-Wei Chang, Kenton Lee, and Kristina Toutanova. 2019.
\newblock Bert: Pre-training of deep bidirectional transformers for language understanding.
\newblock In \emph{Proceedings of the 2019 Conference of the North American Chapter of the Association for Computational Linguistics: Human Language Technologies, Volume 1 (Long and Short Papers)}, pages 4171--4186.

\bibitem[{Dima et~al.(2024)Dima, Avram, Craciun, and Cercel}]{dima2024roqllama}
George-Andrei Dima, Andrei-Marius Avram, Cristian-George Craciun, and Dumitru-Clementin Cercel. 2024.
\newblock Roqllama: A lightweight romanian adapted language model.
\newblock In \emph{Findings of the Association for Computational Linguistics: EMNLP 2024}, pages 4531--4541.

\bibitem[{Dong et~al.(2018)Dong, Shen, Crawford, van Hoof, and Cheung}]{dong2018banditsum}
Yue Dong, Yikang Shen, Eric Crawford, Herke van Hoof, and Jackie Chi~Kit Cheung. 2018.
\newblock Banditsum: Extractive summarization as a contextual bandit.
\newblock In \emph{Proceedings of the 2018 Conference on Empirical Methods in Natural Language Processing}, pages 3739--3748.

\bibitem[{Dubey et~al.(2024)Dubey, Jauhri, Pandey, Kadian, Al-Dahle, Letman, Mathur, Schelten, Yang, Fan et~al.}]{dubey2024llama}
Abhimanyu Dubey, Abhinav Jauhri, Abhinav Pandey, Abhishek Kadian, Ahmad Al-Dahle, Aiesha Letman, Akhil Mathur, Alan Schelten, Amy Yang, Angela Fan, et~al. 2024.
\newblock The llama 3 herd of models.
\newblock \emph{arXiv preprint arXiv:2407.21783}.

\bibitem[{Dumitrescu et~al.(2021)Dumitrescu, Rebeja, Lorincz, Gaman, Avram, Ilie, Pruteanu, Stan, Rosia, Iacobescu, Morogan, Dima, Marchidan, Rebedea, Chitez, Yogatama, Ruder, Ionescu, Pascanu, and Patraucean}]{dumitrescu_nips}
Stefan Dumitrescu, Petru Rebeja, Beata Lorincz, Mihaela Gaman, Andrei Avram, Mihai Ilie, Andrei Pruteanu, Adriana Stan, Lorena Rosia, Cristina Iacobescu, Luciana Morogan, George Dima, Gabriel Marchidan, Traian Rebedea, Madalina Chitez, Dani Yogatama, Sebastian Ruder, Radu~Tudor Ionescu, Razvan Pascanu, and Viorica Patraucean. 2021.
\newblock Liro: Benchmark and leaderboard for romanian language tasks.
\newblock In \emph{Proceedings of the Neural Information Processing Systems Track on Datasets and Benchmarks}, volume~1.

\bibitem[{Dumitrescu and Avram(2020)}]{dumitrescu2020introducing}
{\c{S}}tefan~Daniel Dumitrescu and Andrei-Marius Avram. 2020.
\newblock Introducing ronec-the romanian named entity corpus.
\newblock In \emph{Proceedings of the Twelfth Language Resources and Evaluation Conference}, pages 4436--4443.

\bibitem[{Echim et~al.(2023)Echim, Sm{\u{a}}du, Avram, Cercel, and Pop}]{echim2023adversarial}
Sebastian-Vasile Echim, R{\u{a}}zvan-Alexandru Sm{\u{a}}du, Andrei-Marius Avram, Dumitru-Clementin Cercel, and Florin Pop. 2023.
\newblock Adversarial capsule networks for romanian satire detection and sentiment analysis.
\newblock In \emph{International Conference on Applications of Natural Language to Information Systems}, pages 428--442. Springer.

\bibitem[{Eddine et~al.(2021)Eddine, Tixier, and Vazirgiannis}]{eddine2021barthez}
Moussa~Kamal Eddine, Antoine Tixier, and Michalis Vazirgiannis. 2021.
\newblock Barthez: a skilled pretrained french sequence-to-sequence model.
\newblock In \emph{Proceedings of the 2021 Conference on Empirical Methods in Natural Language Processing}, pages 9369--9390.

\bibitem[{El-Shangiti et~al.(2024)El-Shangiti, Alwajih, and Abdul-Mageed}]{el2024arabic}
Ahmed El-Shangiti, Fakhraddin Alwajih, and Muhammad Abdul-Mageed. 2024.
\newblock Arabic automatic story generation with large language models.
\newblock In \emph{Proceedings of The Second Arabic Natural Language Processing Conference}, pages 140--152.

\bibitem[{Ganin et~al.(2016)Ganin, Ustinova, Ajakan, Germain, Larochelle, Laviolette, March, and Lempitsky}]{ganin2016domain}
Yaroslav Ganin, Evgeniya Ustinova, Hana Ajakan, Pascal Germain, Hugo Larochelle, Fran{\c{c}}ois Laviolette, Mario March, and Victor Lempitsky. 2016.
\newblock Domain-adversarial training of neural networks.
\newblock \emph{Journal of machine learning research}, 17(59):1--35.

\bibitem[{Gonz{\'a}lez et~al.(2022)Gonz{\'a}lez, Louis, and Cheung}]{gonzalez2022source}
Jos{\'e}-{\'A}ngel Gonz{\'a}lez, Annie Louis, and Jackie Chi~Kit Cheung. 2022.
\newblock Source-summary entity aggregation in abstractive summarization.
\newblock In \emph{Proceedings of the 29th International Conference on Computational Linguistics}, pages 6019--6034.

\bibitem[{Hermann et~al.(2015)Hermann, Kocisky, Grefenstette, Espeholt, Kay, Suleyman, and Blunsom}]{hermann2015teaching}
Karl~Moritz Hermann, Tomas Kocisky, Edward Grefenstette, Lasse Espeholt, Will Kay, Mustafa Suleyman, and Phil Blunsom. 2015.
\newblock Teaching machines to read and comprehend.
\newblock \emph{Advances in neural information processing systems}, 28.

\bibitem[{Hoefels et~al.(2022)Hoefels, {\c{C}}{\"o}ltekin, and M{\u{a}}droane}]{hoefels2022coroseof}
Diana~Constantina Hoefels, {\c{C}}a{\u{g}}r{\i} {\c{C}}{\"o}ltekin, and Irina~Diana M{\u{a}}droane. 2022.
\newblock Coroseof-an annotated corpus of romanian sexist and offensive tweets.
\newblock In \emph{Proceedings of the Thirteenth Language Resources and Evaluation Conference}, pages 2269--2281.

\bibitem[{Jiang et~al.(2023)Jiang, Sablayrolles, Mensch, Bamford, Chaplot, Casas, Bressand, Lengyel, Lample, Saulnier et~al.}]{jiang2023mistral}
Albert~Q Jiang, Alexandre Sablayrolles, Arthur Mensch, Chris Bamford, Devendra~Singh Chaplot, Diego de~las Casas, Florian Bressand, Gianna Lengyel, Guillaume Lample, Lucile Saulnier, et~al. 2023.
\newblock Mistral 7b.
\newblock \emph{arXiv preprint arXiv:2310.06825}.

\bibitem[{Jiang and Dreyer(2024)}]{jiang2024ccsum}
Xiang Jiang and Markus Dreyer. 2024.
\newblock Ccsum: A large-scale and high-quality dataset for abstractive news summarization.
\newblock In \emph{Proceedings of the 2024 Conference of the North American Chapter of the Association for Computational Linguistics: Human Language Technologies (Volume 1: Long Papers)}, pages 7299--7329.

\bibitem[{Jitaru et~al.(2020)Jitaru, Abdulamit, and Ionescu}]{jitaru2020lrro}
Andrei~Cosmin Jitaru, {\c{S}}eila Abdulamit, and Bogdan Ionescu. 2020.
\newblock Lrro: a lip reading data set for the under-resourced romanian language.
\newblock In \emph{Proceedings of the 11th ACM Multimedia Systems Conference}, pages 267--272.

\bibitem[{Karim et~al.(2024)Karim, Usama, Ibrahim, Hatem, Wael, Mazrua, El-Monayer, Elbanhawy, Foad, and Moawad}]{karim2024arabic}
Abdelrhman~A Karim, Mohammed Usama, Manar~A Ibrahim, Youssef Hatem, Waleed Wael, Ahmad~T Mazrua, Ghada~K El-Monayer, Magy Elbanhawy, Khaled Foad, and Ibrahim~F Moawad. 2024.
\newblock Arabic abstractive summarization using the multilingual t5 model.
\newblock In \emph{2024 6th International Conference on Computing and Informatics (ICCI)}, pages 223--228. IEEE.

\bibitem[{Kry{\'s}ci{\'n}ski et~al.(2022)Kry{\'s}ci{\'n}ski, Rajani, Agarwal, Xiong, and Radev}]{kryscinski2022booksum}
Wojciech Kry{\'s}ci{\'n}ski, Nazneen Rajani, Divyansh Agarwal, Caiming Xiong, and Dragomir Radev. 2022.
\newblock Booksum: A collection of datasets for long-form narrative summarization.
\newblock In \emph{Findings of the Association for Computational Linguistics: EMNLP 2022}, pages 6536--6558.

\bibitem[{Kwon et~al.(2023)Kwon, Kamigaito, and Okumura}]{kwon2023abstractive}
Jingun Kwon, Hidetaka Kamigaito, and Manabu Okumura. 2023.
\newblock Abstractive document summarization with summary-length prediction.
\newblock In \emph{Findings of the Association for Computational Linguistics: EACL 2023}, pages 618--624.

\bibitem[{Lam et~al.(2023)Lam, Doan, Pham, and Kalita}]{lam2023abstractive}
Khang Lam, Thieu Doan, Khang Pham, and Jugal Kalita. 2023.
\newblock Abstractive text summarization using the brio training paradigm.
\newblock In \emph{Findings of the Association for Computational Linguistics: ACL 2023}, pages 92--99.

\bibitem[{Lewis et~al.(2020)Lewis, Liu, Goyal, Ghazvininejad, Mohamed, Levy, Stoyanov, and Zettlemoyer}]{lewis2020bart}
Mike Lewis, Yinhan Liu, Naman Goyal, Marjan Ghazvininejad, Abdelrahman Mohamed, Omer Levy, Veselin Stoyanov, and Luke Zettlemoyer. 2020.
\newblock Bart: Denoising sequence-to-sequence pre-training for natural language generation, translation, and comprehension.
\newblock In \emph{Proceedings of the 58th Annual Meeting of the Association for Computational Linguistics}, pages 7871--7880.

\bibitem[{Lin(2004)}]{lin2004rouge}
Chin-Yew Lin. 2004.
\newblock Rouge: A package for automatic evaluation of summaries.
\newblock In \emph{Text summarization branches out}, pages 74--81.

\bibitem[{Lin and Hovy(2003)}]{lin2003automatic}
Chin-Yew Lin and Eduard Hovy. 2003.
\newblock Automatic evaluation of summaries using n-gram co-occurrence statistics.
\newblock In \emph{Proceedings of the 2003 human language technology conference of the North American chapter of the association for computational linguistics}, pages 150--157.

\bibitem[{Liu et~al.(2020)Liu, Gu, Goyal, Li, Edunov, Ghazvininejad, Lewis, and Zettlemoyer}]{liu2020multilingual}
Yinhan Liu, Jiatao Gu, Naman Goyal, Xian Li, Sergey Edunov, Marjan Ghazvininejad, Mike Lewis, and Luke Zettlemoyer. 2020.
\newblock Multilingual denoising pre-training for neural machine translation.
\newblock \emph{Transactions of the Association for Computational Linguistics}, 8:726--742.

\bibitem[{M{\u{a}}nescu et~al.(2023)M{\u{a}}nescu, Sm{\u{a}}du, Avram, Cercel, and Pop}]{muanescu2023end}
Emilian-Claudiu M{\u{a}}nescu, R{\u{a}}zvan-Alexandru Sm{\u{a}}du, Andrei-Marius Avram, Dumitru-Clementin Cercel, and Florin Pop. 2023.
\newblock End-to-end lip reading in romanian with cross-lingual domain adaptation and lateral inhibition.
\newblock In \emph{2023 IEEE International Conference on Web Intelligence and Intelligent Agent Technology (WI-IAT)}, pages 287--293. IEEE.

\bibitem[{Mao et~al.(2022)Mao, Zhong, and Han}]{mao2022citesum}
Yuning Mao, Ming Zhong, and Jiawei Han. 2022.
\newblock Citesum: Citation text-guided scientific extreme summarization and domain adaptation with limited supervision.
\newblock In \emph{Proceedings of the 2022 Conference on Empirical Methods in Natural Language Processing}, pages 10922--10935.

\bibitem[{Masala et~al.(2024)Masala, Ilie-Ablachim, Dima, Corlatescu, Zavelca, Olaru, Terian-Dan, Terian-Dan, Leordeanu, Velicu et~al.}]{masala2024vorbe}
Mihai Masala, Denis~C Ilie-Ablachim, Alexandru Dima, Dragos Corlatescu, Miruna Zavelca, Ovio Olaru, Simina Terian-Dan, Andrei Terian-Dan, Marius Leordeanu, Horia Velicu, et~al. 2024.
\newblock " vorbe$\backslash$c $\{$s$\}$ ti rom$\backslash$\^{} ane$\backslash$c $\{$s$\}$ te?" a recipe to train powerful romanian llms with english instructions.
\newblock \emph{arXiv preprint arXiv:2406.18266}.

\bibitem[{Matei et~al.(2024)Matei, T{\u{a}}iatu, Sm{\u{a}}du, and Cercel}]{matei2024enhancing}
Vlad-Cristian Matei, Iulian-Marius T{\u{a}}iatu, R{\u{a}}zvan-Alexandru Sm{\u{a}}du, and Dumitru-Clementin Cercel. 2024.
\newblock Enhancing romanian offensive language detection through knowledge distillation, multi-task learning, and data augmentation.
\newblock In \emph{International Conference on Applications of Natural Language to Information Systems}, pages 317--332. Springer.

\bibitem[{Mendes et~al.(2019)Mendes, Narayan, Miranda, Marinho, Martins, and Cohen}]{mendes2019jointly}
Alfonso Mendes, Shashi Narayan, Sebasti{\~a}o Miranda, Zita Marinho, Andr{\'e}~FT Martins, and Shay~B Cohen. 2019.
\newblock Jointly extracting and compressing documents with summary state representations.
\newblock In \emph{Proceedings of the 2019 Conference of the North American Chapter of the Association for Computational Linguistics: Human Language Technologies, Volume 1 (Long and Short Papers)}, pages 3955--3966.

\bibitem[{Nallapati et~al.(2016)Nallapati, Zhou, dos Santos, Gulcehre, and Xiang}]{nallapati2016abstractive}
Ramesh Nallapati, Bowen Zhou, Cicero dos Santos, Caglar Gulcehre, and Bing Xiang. 2016.
\newblock Abstractive text summarization using sequence-to-sequence rnns and beyond.
\newblock In \emph{Proceedings of the 20th SIGNLL Conference on Computational Natural Language Learning}, pages 280--290.

\bibitem[{Narayan et~al.(2018{\natexlab{a}})Narayan, Cohen, and Lapata}]{narayan2018don}
Shashi Narayan, Shay~B Cohen, and Mirella Lapata. 2018{\natexlab{a}}.
\newblock Don’t give me the details, just the summary! topic-aware convolutional neural networks for extreme summarization.
\newblock In \emph{Proceedings of the 2018 Conference on Empirical Methods in Natural Language Processing}, pages 1797--1807.

\bibitem[{Narayan et~al.(2018{\natexlab{b}})Narayan, Cohen, and Lapata}]{narayan2018ranking}
Shashi Narayan, Shay~B Cohen, and Mirella Lapata. 2018{\natexlab{b}}.
\newblock Ranking sentences for extractive summarization with reinforcement learning.
\newblock In \emph{Proceedings of the 2018 Conference of the North American Chapter of the Association for Computational Linguistics: Human Language Technologies, Volume 1 (Long Papers)}, pages 1747--1759.

\bibitem[{Nicola et~al.(2024)Nicola, Cercel, and Pop}]{nicola2024investigating}
Elena-Beatrice Nicola, Dumitru-Clementin Cercel, and Florin Pop. 2024.
\newblock Investigating the impact of semi-supervised methods with data augmentation on offensive language detection in romanian language.
\newblock \emph{Procedia Computer Science}, 246:3447--3456.

\bibitem[{Niculescu et~al.(2022)Niculescu, Ruseti, and Dascalu}]{niculescu2022rosummary}
Mihai~Alexandru Niculescu, Stefan Ruseti, and Mihai Dascalu. 2022.
\newblock Rosummary: Control tokens for romanian news summarization.
\newblock \emph{Algorithms}, 15(12):472.

\bibitem[{Pavel et~al.(2024)Pavel, Ianina, and Malykh}]{pavel2024sumhis}
Tikhonov Pavel, Anastasiya Ianina, and Valentin Malykh. 2024.
\newblock Sumhis: Extractive summarization exploiting hidden structure.
\newblock \emph{arXiv preprint arXiv:2406.08215}.

\bibitem[{Radford et~al.(2019)Radford, Wu, Child, Luan, Amodei, Sutskever et~al.}]{radford2019language}
Alec Radford, Jeffrey Wu, Rewon Child, David Luan, Dario Amodei, Ilya Sutskever, et~al. 2019.
\newblock Language models are unsupervised multitask learners.
\newblock \emph{OpenAI blog}, 1(8):9.

\bibitem[{Raffel et~al.(2020)Raffel, Shazeer, Roberts, Lee, Narang, Matena, Zhou, Li, and Liu}]{raffel2020exploring}
Colin Raffel, Noam Shazeer, Adam Roberts, Katherine Lee, Sharan Narang, Michael Matena, Yanqi Zhou, Wei Li, and Peter~J Liu. 2020.
\newblock Exploring the limits of transfer learning with a unified text-to-text transformer.
\newblock \emph{Journal of machine learning research}, 21(140):1--67.

\bibitem[{Rogoz et~al.(2021)Rogoz, Mihaela, and Ionescu}]{rogoz2021saroco}
Ana-Cristina Rogoz, Gaman Mihaela, and Radu~Tudor Ionescu. 2021.
\newblock Saroco: Detecting satire in a novel romanian corpus of news articles.
\newblock In \emph{Proceedings of the 59th Annual Meeting of the Association for Computational Linguistics and the 11th International Joint Conference on Natural Language Processing (Volume 2: Short Papers)}, pages 1073--1079.

\bibitem[{Salton and Buckley(1988)}]{salton1988term}
Gerard Salton and Christopher Buckley. 1988.
\newblock Term-weighting approaches in automatic text retrieval.
\newblock \emph{Information processing \& management}, 24(5):513--523.

\bibitem[{Savary et~al.(2018)Savary, Candito, Mititelu, Bej{\v{c}}ek, Cap, {\v{C}}{\'e}pl{\"o}, Cordeiro, Eryi{\u{g}}it, Giouli, van Gompel et~al.}]{savary2018parseme}
Agata Savary, Marie Candito, Verginica~Barbu Mititelu, Eduard Bej{\v{c}}ek, Fabienne Cap, Slavom{\'\i}r {\v{C}}{\'e}pl{\"o}, Silvio~Ricardo Cordeiro, G{\"u}l{\c{s}}en~Cebiro{\u{g}}lu Eryi{\u{g}}it, Voula Giouli, Maarten van Gompel, et~al. 2018.
\newblock Parseme multilingual corpus of verbal multiword expressions.
\newblock In \emph{Multiword expressions at length and in depth: Extended papers from the MWE 2017 workshop}.

\bibitem[{Soriano et~al.(2022)Soriano, Ahuir, Hurtado, and Gonz{\'a}lez}]{soriano2022dacsa}
Encarnaci{\'o}n~Segarra Soriano, Vicent Ahuir, Llu{\'\i}s-F Hurtado, and Jos{\'e} Gonz{\'a}lez. 2022.
\newblock Dacsa: A large-scale dataset for automatic summarization of catalan and spanish newspaper articles.
\newblock In \emph{Proceedings of the 2022 Conference of the North American Chapter of the Association for Computational Linguistics: Human Language Technologies}, pages 5931--5943.

\bibitem[{Sutskever et~al.(2014)Sutskever, Vinyals, and Le}]{sutskever2014sequence}
Ilya Sutskever, Oriol Vinyals, and Quoc~V Le. 2014.
\newblock Sequence to sequence learning with neural networks.
\newblock \emph{Advances in neural information processing systems}, 27.

\bibitem[{Team et~al.(2024)Team, Riviere, Pathak, Sessa, Hardin, Bhupatiraju, Hussenot, Mesnard, Shahriari, Ram{\'e} et~al.}]{team2024gemma}
Gemma Team, Morgane Riviere, Shreya Pathak, Pier~Giuseppe Sessa, Cassidy Hardin, Surya Bhupatiraju, L{\'e}onard Hussenot, Thomas Mesnard, Bobak Shahriari, Alexandre Ram{\'e}, et~al. 2024.
\newblock Gemma 2: Improving open language models at a practical size.
\newblock \emph{arXiv preprint arXiv:2408.00118}.

\bibitem[{Thapa et~al.(2023)Thapa, Rauniyar, Shiwakoti, Poudel, Naseem, and Nasim}]{thapa2023nehate}
Surendrabikram Thapa, Kritesh Rauniyar, Shuvam Shiwakoti, Sweta Poudel, Usman Naseem, and Mehwish Nasim. 2023.
\newblock Nehate: Large-scale annotated data shedding light on hate speech in nepali local election discourse.
\newblock In \emph{ECAI 2023}, pages 2346--2353. IOS Press.

\bibitem[{Touvron et~al.(2023)Touvron, Martin, Stone, Albert, Almahairi, Babaei, Bashlykov, Batra, Bhargava, Bhosale et~al.}]{touvron2023llama}
Hugo Touvron, Louis Martin, Kevin Stone, Peter Albert, Amjad Almahairi, Yasmine Babaei, Nikolay Bashlykov, Soumya Batra, Prajjwal Bhargava, Shruti Bhosale, et~al. 2023.
\newblock Llama 2: Open foundation and fine-tuned chat models.
\newblock \emph{arXiv preprint arXiv:2307.09288}.

\bibitem[{Vaswani et~al.(2017)Vaswani, Shazeer, Parmar, Uszkoreit, Jones, Gomez, Kaiser, and Polosukhin}]{vaswani2017attention}
Ashish Vaswani, Noam Shazeer, Niki Parmar, Jakob Uszkoreit, Llion Jones, Aidan~N Gomez, {\L}ukasz Kaiser, and Illia Polosukhin. 2017.
\newblock Attention is all you need.
\newblock \emph{Advances in neural information processing systems}, 30.

\bibitem[{Wang et~al.(2023)Wang, Xie, Du, and Hu}]{wang2023t5}
Mingye Wang, Pan Xie, Yao Du, and Xiaohui Hu. 2023.
\newblock T5-based model for abstractive summarization: A semi-supervised learning approach with consistency loss functions.
\newblock \emph{Applied Sciences}, 13(12):7111.

\bibitem[{Wenzek et~al.(2020)Wenzek, Lachaux, Conneau, Chaudhary, Guzm{\'a}n, Joulin, and Grave}]{wenzek2020ccnet}
Guillaume Wenzek, Marie-Anne Lachaux, Alexis Conneau, Vishrav Chaudhary, Francisco Guzm{\'a}n, Armand Joulin, and {\'E}douard Grave. 2020.
\newblock Ccnet: Extracting high quality monolingual datasets from web crawl data.
\newblock In \emph{Proceedings of the Twelfth Language Resources and Evaluation Conference}, pages 4003--4012.

\bibitem[{Xie et~al.(2022)Xie, Bishop, Tiwari, and Ananiadou}]{xie2022pre}
Qianqian Xie, Jennifer~Amy Bishop, Prayag Tiwari, and Sophia Ananiadou. 2022.
\newblock Pre-trained language models with domain knowledge for biomedical extractive summarization.
\newblock \emph{Knowledge-Based Systems}, 252:109460.

\bibitem[{Yao et~al.(2018)Yao, Zhang, Luo, and Wu}]{yao2018deep}
Kaichun Yao, Libo Zhang, Tiejian Luo, and Yanjun Wu. 2018.
\newblock Deep reinforcement learning for extractive document summarization.
\newblock \emph{Neurocomputing}, 284:52--62.

\bibitem[{Zhang et~al.(2023)Zhang, Liu, and Zhang}]{zhang2023extractive}
Haopeng Zhang, Xiao Liu, and Jiawei Zhang. 2023.
\newblock Extractive summarization via chatgpt for faithful summary generation.
\newblock In \emph{Findings of the Association for Computational Linguistics: EMNLP 2023}, pages 3270--3278.

\bibitem[{Zhang et~al.(2020)Zhang, Zhao, Saleh, and Liu}]{zhang2020pegasus}
Jingqing Zhang, Yao Zhao, Mohammad Saleh, and Peter Liu. 2020.
\newblock Pegasus: Pre-training with extracted gap-sentences for abstractive summarization.
\newblock In \emph{International conference on machine learning}, pages 11328--11339. PMLR.

\bibitem[{Zhong et~al.(2020)Zhong, Liu, Chen, Wang, Qiu, and Huang}]{zhong2020extractive}
Ming Zhong, Pengfei Liu, Yiran Chen, Danqing Wang, Xipeng Qiu, and Xuan-Jing Huang. 2020.
\newblock Extractive summarization as text matching.
\newblock In \emph{Proceedings of the 58th Annual Meeting of the Association for Computational Linguistics}, pages 6197--6208.

\end{thebibliography}

\appendix

\section{Dataset Cleaning Rules}
\label{app:clean_rules}

The most important rules that we used to clean the RoLargeSum dataset were:

\begin{itemize}
    \item Removing news articles that had less than 100 characters.
    \item Removing samples for which the ratio between the news article and its corresponding summary was less than 1.2x. 
    \item Removing news articles that contained more than 15\% of their content written in the Cyrillic alphabet\footnote{Most of the news that used the Cyrillic alphabet were written in Russian and came from the Moldavian websites.}.
    \item Removing crawling artifacts found in news articles such as ``Articolul următor'' (eng. ``Next article'') or ``Citește și'' (eng. ``Read also'').
\end{itemize}

\section{RoLargeSum Statistics}
\label{app:dialect_stats}


We outline the statistics of the RoLargeSum dataset for the news articles written by people in Romania and the Republic of Moldova in Table \ref{tab:dialect_stats_ro} and Table \ref{tab:dialect_stats_md}, respectively.

\begin{table*}[t]
\begin{tabular}{l|c|cc|cc|cc}
     \toprule
     \multirow{ 2}{*}{\textbf{Dataset}} & \multirow{ 2}{*}{\textbf{Train/Val/Test}} & \multicolumn{2}{c|}{\textbf{Avg. News Len.}} & \multicolumn{2}{c|}{\textbf{Avg. Sum. Len.}} & \multicolumn{2}{c}{\textbf{Vocab. Size}} \\
     & & \textbf{Words} & \textbf{Sents.} & \textbf{Words} & \textbf{Sents.} & \textbf{News} & \textbf{Sums.} \\
     \midrule
     RoLargeSum-Summary & 108K/2.5K/2.5K & 387.46 & 12.57 & 49.56 & 1.48 & 496K & 153K \\
     RoLargeSum-Headline & 104K/2.5K/2.5K & 389.12 & 12.69 & 18.73 & 1.40 & 491K & 81K \\
     RoLargeSum-Keywords & 108K/2.5K/2.5K & 387.58 & 12.58 & 9.31 & 1.00 & 496K & 42K \\
     \bottomrule
\end{tabular}
\centering
\caption{RoLargeSum dataset statistics for the news articles from Romania.}
\label{tab:dialect_stats_ro}
\end{table*}

\begin{table*}[t]
\begin{tabular}{l|c|cc|cc|cc}
     \toprule
     \multirow{ 2}{*}{\textbf{Dataset}} & \multirow{ 2}{*}{\textbf{Train/Val/Test}} & \multicolumn{2}{c|}{\textbf{Avg. News Len.}} & \multicolumn{2}{c|}{\textbf{Avg. Sum. Len.}} & \multicolumn{2}{c}{\textbf{Vocab. Size}} \\
     & & \textbf{Words} & \textbf{Sents.} & \textbf{Words} & \textbf{Sents.} & \textbf{News} & \textbf{Sums.} \\
     \midrule
     RoLargeSum-Summary & 411/2.5K/2.5K & 314.78 & 9.08 & 46.80 & 1.63 & 897K & 206K \\
     RoLargeSum-Headline & 499K/2.5K/2.5K & 314.80 & 9.04 & 15.21 & 1.17 & 1.0M & 118K \\
     RoLargeSum-Keywords & 308K/2.5K/2.5K & 313.45 & 7.75 & 6.95 & 1.00 & 774K & 47K \\
     \bottomrule
\end{tabular}
\centering
\caption{RoLargeSum dataset statistics for the news articles from the Republic of Moldova.}
\label{tab:dialect_stats_md}
\end{table*}

\section{TF-IDF Analysis}
\label{app:tfidf}

We perform a TF-IDF analysis of the news articles found in RoLargeSum. The product of TF and ID is known as TF-IDF, where: (i) TF counts the frequency of each word that occurs in a particular document, and (ii) IDF calculates the frequency of the corresponding term in all documents \cite{salton1988term, thapa2023nehate}. We note that the terms that are infrequent in the entire dataset but frequent in a small selection of texts are indicated by higher TF-IDF scores.

The results are summarized in Table \ref{tab:tfidf}. An important observation of this analysis is the domain of the extracted words. The words in the Romanian news articles with the highest TF-IDF score belong to Western company names (e.g., ``Analytica'', ``Kogan'', ``Facebook'') or the technology domain (e.g., ``aplicație'' --- eng. ``application'', ``colectat'' --- eng. ``collected''). Furthermore, most of the terms in the Moldavian news articles with the highest TF-IDF score are based on political names (e.g., ``Voronin'', ``Chetaru''), political office jobs (e.g., ``Președintelui'' --- eng. ``President's'', ``Premierului'' --- eng. ``Prime Minister's''), or political parties (e.g., ``PCRM'' --- eng. ``Party of Communists of the Republic of Moldova'', ``PLDM'' --- eng. ``Liberal Democratic Party of Moldova''). Thus, this analysis might indicate that the 
topic information of Romanian news is more externally focused, mostly on Western media, while Moldavian news is more internally focused on internal politics.

\begin{table*}[!ht]
\centering
\begin{tabular}{l|l|l|c}
    \toprule
         \textbf{Country Name} & \textbf{Words} & \textbf{Translation} & \textbf{TF-IDF Score} \\
         \midrule
         \multirow{10}{*}{\textbf{Romania}}& Analytica & Analytica & 0.447 \\
          & Cambridge & Cambridge & 0.386 \\
          & Kogan & Kogan & 0.301 \\
          & Wylie & Wylie & 0.191 \\
          & Facebook & Moldavia's & 0.176 \\
          & Alexandr & Alexandr & 0.168 \\
          & colectat & collected & 0.132 \\
          & Christopher & Christopher & 0.128 \\
          & aplicație & application & 0.118 \\
          & utilizatorilor & user's & 0.116 \\
          \midrule
         \multirow{10}{*}{\textbf{Moldova}} & Voronin & Voronin & 0.231 \\
          & Chetraru & Chetraru & 0.219 \\
          & Președintelui & President's & 0.209 \\
          & propunerea & the proposal & 0.201 \\
          & Anticorupție & Anti-corruption & 0.181 \\
          & Național & National & 0.153 \\
          & PCRM & PCRM & 0.146 \\
          & Parlamentului & Parliament's & 0.144 \\
          & Premierului & Prime Minister's & 0.142 \\
          & PLDM & LDPM & 0.142 \\
    \bottomrule
\end{tabular}
\caption{TF-IDF analysis of the top ten most common words in the RoLargeSum news articles in Romania and Moldova.}

\label{tab:tfidf}
\end{table*}

\section{Large Language Models}
\label{app:llms}

We evaluate the performance of several LLMs that are pre-trained on Romanian or multilingual datasets and fall into one of three model families: Llama, Gemma, and Mistral.

\paragraph{Llama} We perform experiments with LLama2-7B \cite{touvron2023llama}, LLama3-8B, and LLama3.1-8B \cite{dubey2024llama}, using both the base and the instruction fine-tuned versions for all these variants.

\paragraph{Gemma} We also test the performance of three Gemma LLMS (i.e., Gemma-7B, Gemma-7B-It, and Gemma-1.1.-7B-It) \cite{team2024gemma}, and two Gemma2 variants (i.e., Gemma2-9B and Gemma2-9B-It) \cite{team2024gemma} on RoLargeSum.

\paragraph{Mistral} We further evaluate the performance of the 7B Mistral versions (i.e., v0.1, v0.2, and v0.3) \cite{jiang2023mistral}, using both the base and the instruction fine-tuned variants.

\paragraph{Romanian LLMs} 
Finally, we assess the performance of Romanian LLMs on RoLargeSum, which belong to the three families of LLMs that we were interested in as follows: RoLLama2-7B, RoLlama2-7B-It, RoLlama3-8B-It, RoMistral-7B-It, RoGemma-7B-It \cite{masala2024vorbe}, and RoQLlama-7B \cite{dima2024roqllama}.


\section{Evaluation Metrics}
\label{app:rouge}

Three versions of the Recall-Oriented Understudy for Gisting Evaluation (ROUGE) score \cite{lin2003automatic} were used to assess all of the models we used in our experiments: unigram-based ROUGE (R-1), bigram-based ROUGE (R-2), and longest common sequence ROUGE (R-L). Equation \ref{eq:rouge_1} provides the formula used to determine the R-1 and R-2 scores:
\begin{equation}
    \mathrm{R-N} = \frac{\sum\limits_{S \in \mathcal{D}}\sum\limits_{g \in S} \mathrm{Count}_{\mathrm{match}} (g)}{\sum\limits_{S \in \mathcal{D}}\sum\limits_{g \in S} \mathrm{Count} (g)}
    \label{eq:rouge_1}
\end{equation}
where $\mathcal{D}$ is the dataset containing the reference summaries, $S$ is a summary in this dataset, $g$ is either a unigram or bigram in $S$, and $\mathrm{Count}_{\mathrm{match}}$ calculates the matching grams between the reference summary $S$ and the candidate summary, and $\mathrm{Count}$ simply counts the grams found in $S$.

On the other hand, R-L is computed as described in Equation \ref{eq:rouge_lcs}:
\begin{equation}
    \mathrm{R-L} = \frac{LCS(C, S)}{|S|}
    \label{eq:rouge_lcs}
\end{equation}
where $LCS$ is the longest common sequence between the candidate summary $C$ and the reference summary $S$, and $|S|$ represents the word count in $S$. 

\section{Implementation Details}
\label{app:implem}

We performed a linear warm-up for the first 5k steps of the training process and fine-tuned the BART models for ten epochs with a learning rate of 3e-5, a batch size of 16, and a gradient accumulation of 4 for a total batch size of 64. During inference, we decoded the output using a beam search with a beam width of 16. For adversarial training, we set $\lambda_{GR}=0.01$ for the gradient reversal method and $\lambda_{LR}=0.005$ for the loss reversal method. 

For the summaries generation task using Romanian LLMs, we simply used HuggingingFace's pipeline for text generation with the default parameters\footnote{\url{https://huggingface.co/docs/transformers/en/main_classes/pipelines}}. 

\section{Summarization Prompt}
\label{app:prompt}

We evaluated the LLMs on the RoLargeSum dataset for Romanian summarization by inserting both the headline of the article and the text contained in the article using the following prompt:

\begin{queryl}[linewidth=7.5cm]
Eu citesc urmatorul paragraf si 
il sumarizez.

Titlu: {titlu}

Text: {text}

Sumarizare:
\end{queryl}
where the fields in the curly brackets are filled with the corresponding headlines and documents in the test set of RoLargeSum.

The English version of the prompt is the following:

\begin{queryl}[linewidth=7.5cm]
I read the following paragraph 
and summarize it.

Headline: {headline}

Text: {Text}

Summary:
\end{queryl}

\section{Generation Examples}
\label{app:examples}

In Tables \ref{tab:summ_examples_1}, \ref{tab:summ_examples_2} and \ref{tab:summ_examples_3}, we show several examples that were generated using the best-performing system (i.e., mBART-large with Unlimiformer and Loss Reversal) for summaries, headlines, and keywords, both in Romanian and translated into English. We can observe that the generated summaries, headlines, and keywords align with their respective reference text and the original news article. 

\paragraph{First Example} In the first example, the model does not recall in the summary the information on the dissolution of the federal agency that identified wildfire outbreaks in Siberia presented in the last part of the article. The generated headline fits well with the main subject of the article. The model was able to extract several relevant keywords, although they were not identical to the reference keywords.

\paragraph{Second Example} In the second example, the top-performing system was able to extract the main takeaways of the article in the summary. It generated an almost identical headline to the reference headline and correctly identified all the reference keywords except the first one.

\paragraph{Third Example} In the third example, the model does not incorporate the summary of the information presented in the last part of the article on the employment status on the Romanian coast. The candidate headline fits well with the main message of the article and is much shorter than the candidate headline. The extracted keywords are relevant to the article, but the model misses an important one about the Mamaia resort; however, it correctly identifies the Vama Veche resort. Also, it captures an important keyword (i.e., ``salarii'' --- eng. ``salaries'') that is absent from the reference keywords.

\begin{table*}[!ht]
    \centering
    \resizebox{!}{.4\paperheight}{
    \begin{tabular}{|p{16.1cm}|}
         \toprule
         \textbf{Document:} Raportul ONU pe urgența schimbărilor climatice e confirmat de o informație de la NASA. Pentru prima dată în istorie, fumul de la incendiile de vegetație din Siberia a ajuns la Polul Nord. Nici nu e de mirare. În regiunea Yakutia ard 3,4 milioane de hectare de pădure, iarbă și arbuști. Într-un sat, 400 de locuitori au fost evacuați sâmbătă din calea flăcărilor. Acum, ei s-au întors dar au găsit totul în ruine. „Au ars case multe. 34 de case, e un număr mare. Erau multe apartamente. Două case cu câte patru apartamente au ars pe partea asta. Două case cu câte două apartamente au ars în partea cealaltă. Au ars și case private”, a spus Ilya Avakumov, locuitor din Byas-Kuel.Incendiile din Siberia sunt provocate de verile tot mai calde. La proporțiile dezastrului a contribuit și statul, care a desființat, în 2007, o agenție aviatică federală care avea ca scop identificarea focarelor cu ajutorul avioanelor. În plus, o lege permite autorităților locale să lase incendiile să ardă, dacă flăcările nu amenință satele, iar costurile de stingere sunt prea mari. \\ \\
         \textbf{Reference Summary:} Fumul de la incendiile din Siberia a ajuns la Polul Nord, la 3.000 de km depărtare. Anul acesta, focul a distrus peste 14 milioane de hectare de vegetație în Siberia. O agenție care repera din timp focarele a fost desființată în 2007. \\
         \textbf{Candidate Summary:} Pentru prima dată în istorie, incendiile de vegetație din Siberia au ajuns la Polul Nord. În regiunea Yakutia ard 3,4 milioane de hectare de pădure, iarbă și arbuști. Incendiile din Siberia sunt provocate de verile tot mai calde. \\ \\

         \textbf{Reference Headline:} Fumul de la incendiile din Siberia a ajuns la Polul Nord. Focul a distrus deja 14 milioane de hectare de vegetație \\
         \textbf{Candidate Headline:} Pentru prima dată în istorie, fumul de la incendiile de vegetație din Siberia a ajuns la Polul Nord \\ \\

         \textbf{Reference Keywords:} incendii siberia, polul nord, schimbri climatice \\
         \textbf{Candidate Keywords:} siberia, incendii de vegetatie, polul nord \\

         \midrule
         \textbf{Document:} The UN report on the urgency of climate change is confirmed by information from NASA. For the first time in history, the smoke from wildfires in Siberia has reached the North Pole. It is no wonder, either. In the Yakutia region, 3.4 million hectares of forest, grass and shrubs are burning. In one village, 400 inhabitants were evacuated on Saturday from the path of the flames. Now, they returned but found everything in ruins. "Many houses burned down. Thirty-four houses, that's a big number. There were many apartments. Two houses with four apartments each burned on this side. Two houses with two apartments each burned on the other side. Private houses also burned," said Ilya Avakumov, a resident of Byas-Kuel. The fires in Siberia are caused by increasingly hot summers. The state also contributed to the scale of the disaster, which in 2007 abolished a federal aviation agency that aimed to identify outbreaks with the help of aeroplanes. In addition, a law allows local authorities to let the fires burn if the flames do not threaten villages and the costs of extinguishing them are too high. \\ \\

         \textbf{Reference Summary:} Smoke from fires in Siberia reached the North Pole, 3,000 km away. This year, the fire destroyed more than 14 million hectares of vegetation in Siberia. An agency that used to spot outbreaks early was disbanded in 2007. \\
         \textbf{Candidate Summary:} For the first time in history, wildfires in Siberia have reached the North Pole. In the Yakutia region, 3.4 million hectares of forest, grass and shrubs are burning. The fires in Siberia are caused by increasingly hot summers. \\ \\

         \textbf{Reference Headline:} Smoke from fires in Siberia has reached the North Pole. The fire has already destroyed 14 million hectares of vegetation \\
         \textbf{Candidate Headline:} For the first time in history, smoke from wildfires in Siberia has reached the North Pole \\ \\

         \textbf{Reference Keywords:} wildfires in siberia, the north pole, climate change \\
         \textbf{Candidate Keywords:} siberia, vegetation wildfires, the north pole \\
         
         \bottomrule
    \end{tabular}
    }
    \caption{The first example of generated summaries, headlines and keywords, both in Romanian (top) and translated into English (bottom).}
    \label{tab:summ_examples_1}
\end{table*}

\begin{table*}[!ht]
    \centering
    \resizebox{!}{.4\paperheight}{
    \begin{tabular}{|p{13.73cm}|}
         \toprule
         \textbf{Document:} Meteorologii anunță că vremea se va încălzi în majoritatea regiunilor, dar va fi în general instabilă. În jumătatea de est a țării și în zonele montane vor fi perioade cu instabilitate atmosferică accentuată ce se va manifesta prin averse torențiale, descărcări electrice, intensificări ale vântului și izolat grindină și vijelii. În intervale scurte de timp sau prin acumulare, cantitățile de apă vor depăși pe arii restrânse 25…50 l/mp. În restul teritoriului, cerul va fi variabil, iar astfel de fenomene se vor semnala pe spații mici. Temperaturile maxime se vor încadra între 19 grade pe litoral și 28 de grade Celsius în Câmpia de Vest, iar cele minime între 10 și 17 grade Celsius. Izolat vor fi condiții de ceață. \\ \\

         \textbf{Reference Summary:} Temperaturile vor fi mai crescute, dar va continua să plouă în majoritatea regiunilor. În sudul țării, vremea va fi instabilă. În Capitală, în a doua parte a zilei și la noapte vor fi averse. \\
         \textbf{Candidate Summary:} Meteorologii anunță că vremea se va încălzi în majoritatea regiunilor, dar va fi în general instabilă. Temperaturile maxime se vor încadra între 19 grade pe litoral și 28 de grade Celsius în Câmpia de Vest. Temperaturile minime între 10 și 17 grade Celsius. \\
         
         \textbf{Reference Headline:} Vremea se va încălzi în majoritatea regiunilor, dar va fi în general instabilă \\
         \textbf{Candidate Headline:} Vremea se încălzește în majoritatea regiunilor, dar va fi în general instabilă \\

         \textbf{Reference Keywords:} anm, meteo, prognoza meteo, vreme \\
         \textbf{Candidate Keywords:} meteo, prognoza meteo, vreme \\

         \midrule
         \textbf{Document:} Meteorologists say the weather will warm up in most regions, but it will be generally unsettled. In the eastern half of the country and the mountainous areas, periods of increased atmospheric instability will manifest in which torrential downpours, electrical discharges, wind intensifications, isolated hail, and storms will manifest. In short periods or through accumulation, the amounts of water will exceed 25...50 l/m2 in limited areas. The sky will be variable in the rest of the territory, and such phenomena will be signalled in small spaces. The maximum temperatures will be between 19 degrees on the coast and 28 degrees Celsius in the Western Plain, and the minimum between 10 and 17 degrees Celsius. There will be isolated foggy conditions.  \\ \\

         \textbf{Reference Summary:} The temperatures will be higher, but it will continue to rain in most regions. In the south of the country, the weather will be unstable. In the capital, showers will be provided during the second part of the day and at night. \\
         \textbf{Candidate Summary:} Meteorologists announce that the weather will warm up in most regions but will be generally unstable. Maximum temperatures will be between 19 degrees on the coast and 28 degrees Celsius in the Western Plains. Minimum temperatures between 10 and 17 degrees Celsius. \\
         
         \textbf{Reference Headline:} The weather will warm in most regions, but it will be generally unsettled \\
         \textbf{Candidate Headline:} The weather is warming up in most regions, but will be generally unsettled \\

         \textbf{Reference Keywords:} anm, weather, weather forecast, weather \\
         \textbf{Candidate Keywords:} weather, weather forecast, weather \\
         \bottomrule
    \end{tabular}
    }
    \caption{The second example of generated summaries, headlines and keywords, both in Romanian (top) and translated into English (bottom).}
    \label{tab:summ_examples_2}
\end{table*}

\begin{table*}[!ht]
    \centering
    \resizebox{!}{.4\paperheight}{
    \begin{tabular}{|p{15.61cm}|}
         \toprule
         \textbf{Document:} Patronii de pe litoralul românesc n-au mai apelat anul acesta la forța de muncă străină. Angajatorii spun că au primit zeci de solicitări de a angaja românii întorși acasă din cauza pandemiei. Administratorii din Vama Veche spun că salariile angajaților calificați se duc spre 2.000 de euro, iar următoarea categorie trece de 1.000 de euro. La Mamaia, salariile sunt însă altele. Un barmen câștigă pe lună 3.500-4.000 de lei, iar un ajutor de barman sau ospătar este plătit cu 1.500 -2.500 de lei. Cristian Bălan, administrator terasă Mamaia: „Față de alți ani când aduceam din Republică, aduceam din Bulgaria, găsim personal local. Avem de unde selecta față de alți ani unde mai pe vechi aveai pe birou 50 de CV-uri, acum până în pandemie ne descurcam cu 5-6, iar au apărut solicitările de muncă. „Cornel Lazăr, administrator restaurant-tarasă Vama Veche:„Dacă în alți ani înainte de deschidere căutam disperați angajați, ajunsesem și la soluții de a aduce angajați din afara țării, 2020 a venit cu un surplus de forță de muncă. E primul an în ultimii 10 în care primim oferte de muncă de la lucrători români.” \\ \\

         \textbf{Reference Summary:} Patronii de pe litoralul românesc au zilnic cereri de a angaja români reveniți în țară. Salariile pe litoral încep de la 1.500-2.500 de lei, dar pot ajunge și până la 2.000 de euro. „E primul an în ultimii 10 în care primim oferte de muncă de la lucrători români”, spune administratorul unei terase. \\
         \textbf{Candidate Summary:} Angajatorii spun că au primit zeci de solicitări de a angaja românii întorși acasă din cauza pandemiei. Administratorii din Vama Veche spun că salariile angajaților calificați se duc spre 2.000 de euro. La Mamaia, salariile sunt însă altele. \\
         
         \textbf{Reference Headline:} Latura „bună” a pandemiei. În lipsa sezonierilor străini, patronii de pe litoralul românesc angajează români \\
         \textbf{Candidate Headline:} Patronii de pe litoralul românesc n-au mai apelat anul acesta la forța de muncă străină \\

         \textbf{Reference Keywords:} mamaia, romani, sezonieri straini,vama veche \\
         \textbf{Candidate Keywords:} angajati, angajati straini, salarii, vama veche \\

         \midrule
         \textbf{Document:} Employers on the Romanian coast did not call on foreign labor this year. Employers say they have received dozens of requests to hire Romanians who have returned home due to the pandemic. The administrators of Vama Veche say that the salaries of qualified employees go up to 2,000 euros, and the next category exceeds 1,000 euros. In Mamaia, however, salaries are different. A bartender earns 3,500-4,000 lei per month, and a helper as a bartender or waiter is paid 1,500-2,500 lei. Cristian Bălan, Mamaia terrace administrator: "Compared to other years when we brought from the Republic, we brought from Bulgaria, we find local staff. We have something to choose from compared to other years where, in the old days, you had 50 CVs on your desk. Now, until the pandemic, we managed with 5-6, and the job requests appeared. "Cornel Lazăr, restaurant-terrace administrator Vama Veche: "If in other years before the opening we were desperately looking for employees, we had also reached solutions to bring in employees from outside the country, 2020 came with a labor surplus. It's the first year in the last ten that we receive job offers from Romanian workers." \\ \\

         \textbf{Reference Summary:} Employers on the Romanian coast have daily requests to hire Romanians who have returned to the country. Salaries on the coast start from 1,500-2,500 lei, but can reach up to 2,000 euros. "It's the first year in the last ten that we receive job offers from Romanian workers," says the administrator of a terrace. \\
         \textbf{Candidate Summary:} Employers say that they have received dozens of requests to hire Romanians who have returned home due to the pandemic. The administrators of Vama Veche say that the salaries of qualified employees go up to 2,000 euros. In Mamaia, however, the salaries are different. \\
         
         \textbf{Reference Headline:} The "good" side of the pandemic. In the absence of foreign seasonal workers, employers on the Romanian coast hire Romanians \\
         \textbf{Candidate Headline:} Employers on the Romanian coast did not call on foreign labor this year \\

         \textbf{Reference Keywords:} mamaia, Romanians, seasonal foreigners,vama veche \\
         \textbf{Candidate Keywords:} employees, foreign employees, salaries, vama veche \\
         \bottomrule
    \end{tabular}
    }
    \caption{The third example of generated summaries, headlines, and keywords, both in Romanian (top) and translated into English (bottom).}
    \label{tab:summ_examples_3}
\end{table*}

\end{document}